\documentclass[10pt,twocolumn,letterpaper]{article}

\newif\ifarxiv\arxivtrue

\usepackage[pagenumbers]{cvpr}  

\definecolor{cvprblue}{rgb}{0.21,0.49,0.74}

\usepackage{graphicx}
\usepackage{amsmath}
\usepackage{amssymb}
\usepackage{booktabs}
\usepackage{times}
\usepackage{epsfig}
\usepackage{multirow}
\usepackage{float}
\usepackage{threeparttable}
\usepackage{bbm}
\usepackage{arydshln}
\usepackage{colortbl}
\usepackage[linesnumbered, ruled, lined, boxed, commentsnumbered]{algorithm2e}

\usepackage{amsmath,amssymb,amsthm,bm}
\usepackage{microtype}

\usepackage{makecell}
\usepackage{caption}
\usepackage{tabularx}

\usepackage[pagebackref,breaklinks,colorlinks,allcolors=cvprblue]{hyperref}

\usepackage[capitalize]{cleveref}
\Crefname{section}{Section}{Sections}
\Crefname{table}{Table}{Tables}

\def\paperID{} 
\def\confName{CVPR}
\def\confYear{2026}

\title{Why Not Hyperparameter-Friendly Optimisation? A Monotonic Adaptive Norm Rescaling Approach For Long-Tailed Recognition}

\author{Shuo Zhang\thanks{Corresponding author}\,\,,\, Chenqi Li,\, Tingting Zhu\\
University of Oxford\\
{\tt\small \{shuo.zhang, chenqi.li, tingting.zhu\}@eng.ox.ac.uk}
}

\begin{document}
\maketitle

\begin{abstract}
Long-tailed recognition poses a significant challenge for deep learning. The two-stage decoupling paradigm, which separates representation learning from classifier retraining, offers a promising solution. During the classifier retraining stage, adaptive norm rescaling is a popular technique. It adjusts the per-class weight norms via parameter regularization, which inevitably introduces hyperparameters. However, many studies report that long-tailed recognition is sensitive to these hyperparameters, as their setup significantly impacts performance. In this paper, we first provide a class-conditional distribution perspective to support norm rescaling methods. Furthermore, we propose a simple but effective approach called Self-Adaptive Monotonic Normalization (SAMN). SAMN avoids the need for parameter regularization. It directly enforces monotonicity on per-class weight norms using the Pool Adjacent Violators Algorithm, making the method hyperparameter-friendly. SAMN is a universal strategy that integrates seamlessly with other methods for enhanced performance. Experiments on benchmark datasets demonstrate that our method significantly boosts long-tailed recognition performance, often achieving state-of-the-art results. Codes are available at https://github.com/Zhangshuojackpot/SAMN.
\end{abstract}
\section{Introduction}
\label{sec:intro}

\begin{figure}[!t]
  \centering
  \begin{subfigure}{0.98\linewidth}
    \includegraphics[width=\linewidth]{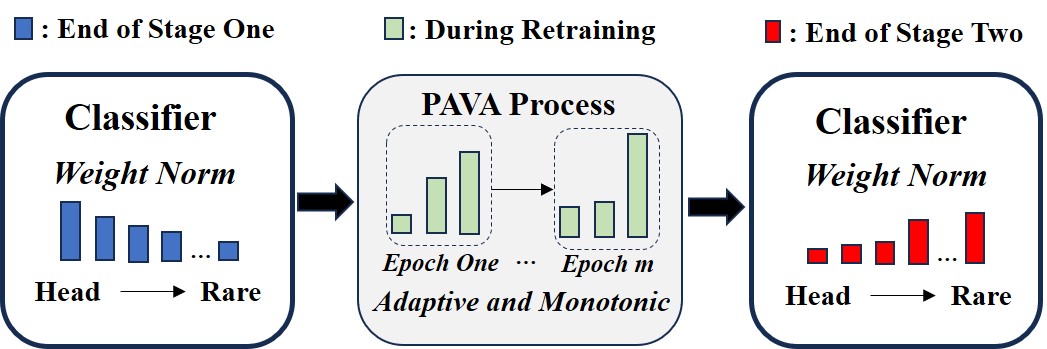}
    \caption{Diagram of Self-Adaptive Monotonic Normalization}
    \label{fig:1a}
  \end{subfigure}
  
  \begin{subfigure}{0.46\linewidth}
    \includegraphics[width=\linewidth]{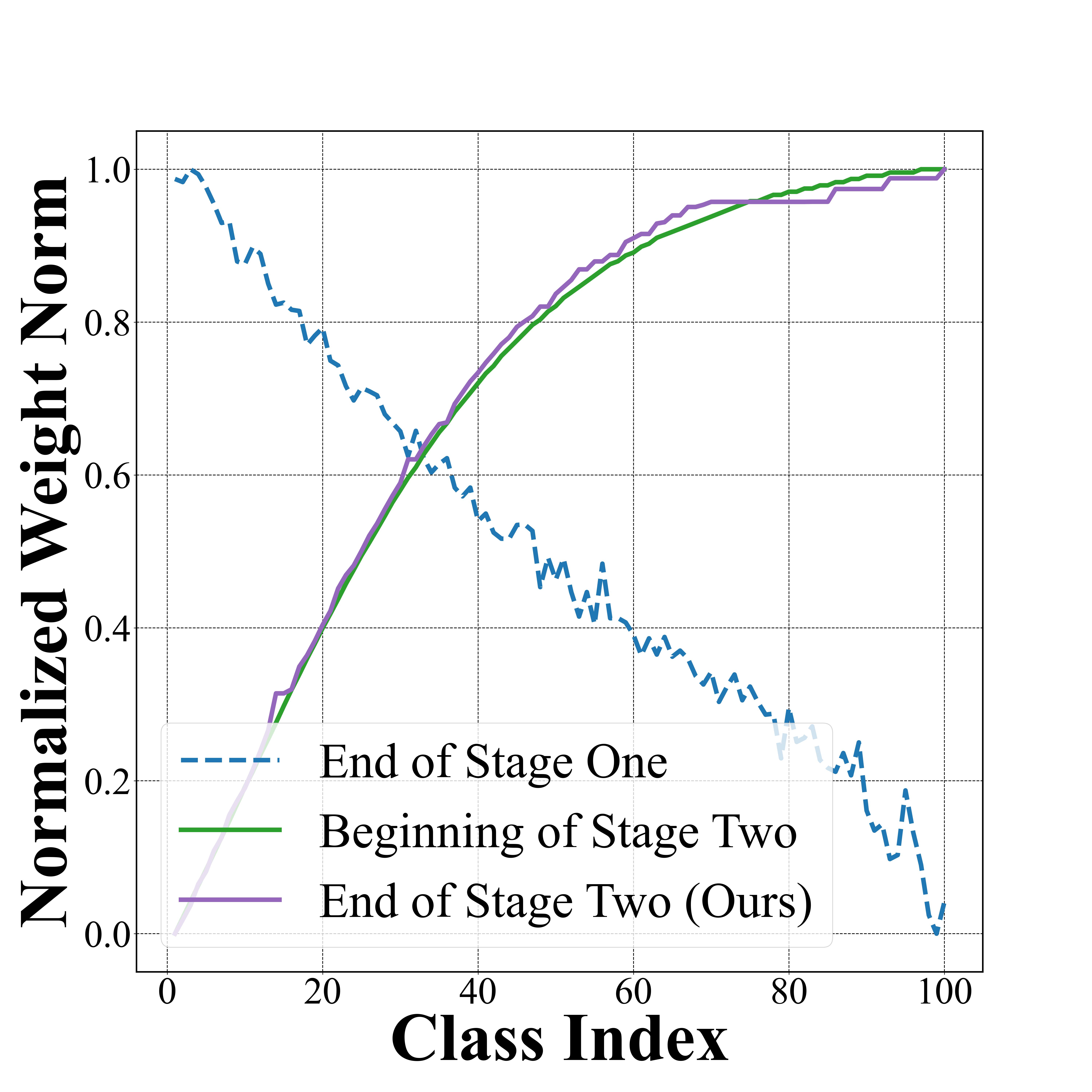}
    \caption{Per-class weight norms changes during training}
    \label{fig:1b}
  \end{subfigure}
  \hfill
  \begin{subfigure}{0.46\linewidth}
    \includegraphics[width=\linewidth]{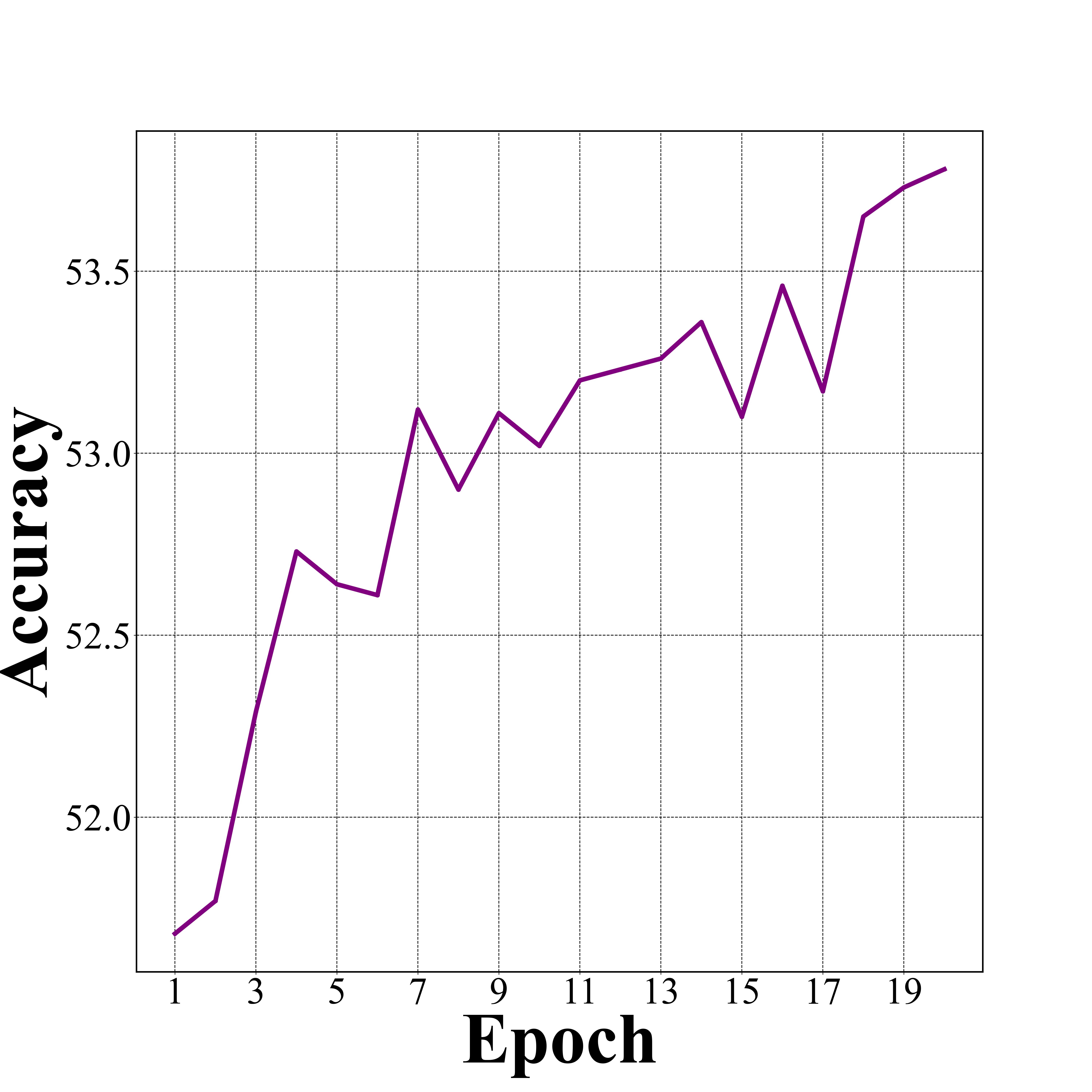}
    \caption{Accuracy (\%) curve during retraining}
    \label{fig:1c}
  \end{subfigure}

  \caption{Diagram and effects of Self-Adaptive Monotonic Normalization (SAMN). We introduce the Pool Adjacent Violators Algorithm (PAVA) to enforce a monotonic ordering of weight norms in a hyperparameter-friendly manner. As shown in (b) and (c), applying SAMN to adjust per-class weight norms leads to further accuracy improvements when the classifier is retrained in the second stage.}
  \label{fig:1}
\end{figure}

Despite their success on balanced datasets, Deep Neural Networks (DNNs) suffer on real-world long-tailed distributions, producing biased results favoring abundant head classes \cite{Cui_2019_CVPR,09217,10105457}. To tackle this, Long-Tailed Recognition (LTR) approaches have emerged, with approaches including: 1) Data rebalancing \cite{10.5555/1622407.1622416,Park_2022_CVPR,10.1007/978-3-319-71249-9_46,Li_2021_CVPR,10.1007/978-3-030-58526-6_41,NEURIPS2023_eeffa70b,ahn2023cuda,shao2024diffultmakediffusionmodel,4717268,JMLR:v18:16-365};
2) Class-balanced loss design \cite{NEURIPS2022_8f4d70db,Du_2023_CVPR,NEURIPS2020_2ba61cc3,Li_2022_CVPR,Cui_2019_CVPR,Lin_2017_ICCV}; 3) Two-stage decoupling \cite{09217,Zhong_2021_CVPR,Alshammari_2022_CVPR,10105458,Du_2023_CVPR,Dang_Yang_Dong_Li_Shi_2024}; and 4) Model ensembling \cite{Zhou_2020_CVPR,wang2021longtailed,NEURIPS2022_dc6319dd}.
The two-stage decoupling strategy has consistently demonstrated remarkable capability and is now regarded as a foundational paradigm in LTR. Consequently, it is the main focus of this paper.

Decoupling strategies divide training into representation learning and classifier retraining. A central theme of the latter is \textit{rescaling per-class weight norms}. The pioneering work \cite{09217} adjusts norms via $\tau$-normalization (using $L_2$-normalization and a margin hyperparameter $\tau$) and Learnable Weight Scaling. \cite{Alshammari_2022_CVPR} observed that naive classifiers yield larger head-class norms, attempting to balance them with techniques like $L_2$-normalization, weight decay, and MaxNorm, while \cite{10105458} applied Class-Balanced Regularization using prior class frequencies as penalty coefficients. Furthermore, \cite{Zhong_2021_CVPR} utilized label smoothing, noting that data mixup benefits representation learning but harms classifier retraining. Recent works include improving lowest recall via Geometric Mean Loss \cite{Du_2023_CVPR} and inverse weight-balancing to compensate for encoder-classifier imbalances \cite{Dang_Yang_Dong_Li_Shi_2024}. These successes highlight that regulating per-class norms is a highly valuable direction for LTR.

Current norm rescaling methods consistently observe that the weight norms of rare classes are considerably smaller than those of head classes \cite{Alshammari_2022_CVPR,10105458,Dang_Yang_Dong_Li_Shi_2024}. Although this observation suggests potential representation underfitting in rare classes, the existing literature reports conflicting interpretations of the underlying cause. 
Some works posit that \textit{underfitting happens in the rare classes}, causing poor performance \cite{Alshammari_2022_CVPR,10105458,Dang_Yang_Dong_Li_Shi_2024,10105459,LI20181}. In contrast, other studies argue that \textit{overfitting happens in the rare classes} \cite{Ye_2021_ICCV,7860349,DBLP:journals/corr/abs-2001-01385,9081988}. 
This fundamental disagreement highlights the need for new evidence and a clear theoretical perspective to resolve the underlying cause of LTR performance degradation. Furthermore, a practical challenge persists: most existing norm rescaling methods rely on parameter regularization, which introduces hyperparameters into the training process. LTR is notoriously sensitive to these hyperparameters, which requires careful and often case-specific tuning across various situations \cite{09217,Zhong_2021_CVPR,Alshammari_2022_CVPR,10105458,Dang_Yang_Dong_Li_Shi_2024,Du_2023_CVPR}.
To address these problems, we first theoretically discuss representation underfitting and overfitting in rare classes from a class-conditional distribution perspective, which underpins the norm rescaling method. Then, we introduce a new norm rescaling method called the Self-Adaptive Monotonic Normalization (SAMN). SAMN is a hyperparameter-friendly method that eliminates parameter regularization by directly enforcing monotonicity of per-class weight norms using the Pool Adjacent Violators Algorithm (PAVA) \cite{ff03617f-3b30-3cb3-a0fc-02979e761de5}. 
As shown in \cref{fig:1a}, we incorporate the PAVA into the classifier retraining stage. This integration adaptively enforces a monotonic constraint: the per-class weight norm remains non-diminishing as an order metric sequence (such as the class frequency) decreases (shown in \cref{fig:1b}). This constraint effectively balances the weights and leads to improved results during the retraining stage (shown in \cref{fig:1c}).
Our contributions can be summarized as follows:
\begin{itemize}
\item {We provide insight into the class-conditional distribution, explaining why poor LTR performance is primarily caused by underfitting rare classes in representation, rather than overfitting them. Our analysis strengthens the foundation of norm rescaling methods.
}
\item {We propose Self-Adaptive Monotonic Normalization (SAMN), a simple yet effective hyperparameter-friendly norm rescaling method. Unlike existing adaptive methods that require sensitive tuning, SAMN offers exceptional convenience and robustness in various scenarios.}
\item{We demonstrate that SAMN is a universal strategy that can be integrated with other LTR techniques to improve performance. Experiments on benchmark and real-world datasets confirm the efficacy of our proposed work.}
\end{itemize}
\section{Related Work}
\label{sec:related_work}

LTR aims to improve the performance of models trained on inherently imbalanced datasets. Methods in this field primarily pursue four main directions:

\noindent\textbf{1) Data Balancing} This classical approach seeks to achieve a more uniform distribution across classes by modifying the dataset sampling strategy. Existing methods either over-sample the input data to ensure the balance of a training batch or produce samples from rare classes using data augmentation strategies \cite{10.5555/1622407.1622416,Park_2022_CVPR,10.1007/978-3-319-71249-9_46,Li_2021_CVPR,10.1007/978-3-030-58526-6_41,NEURIPS2023_eeffa70b,ahn2023cuda,shao2024diffultmakediffusionmodel}.
Alternatively, some techniques achieve balance by undersampling data from head classes \cite{4717268,JMLR:v18:16-365}. 

\noindent\textbf{2) Class-Balanced Loss Design} This direction involves revising the standard loss function to effectively re-weight the contribution of samples or classes during training. The goal is to ensure that rare-class samples have a proper influence on the optimization of the model.
Existing methods assign weights to different classes \cite{Cui_2019_CVPR} or even to an individual sample \cite{Lin_2017_ICCV} to mitigate the dominance of abundant classes.
Furthermore, specific loss formulations have been developed to intrinsically benefit rare-class recognition \cite{NEURIPS2022_8f4d70db,Du_2023_CVPR,NEURIPS2020_2ba61cc3,Li_2022_CVPR}.
These losses often encourage larger decision margins for rare classes relative to more frequent head classes.

\noindent\textbf{3) Two-Stage Decoupling} Methods based on the decoupling paradigm separate the training process into distinct representation learning and classifier retraining stages. A detailed discussion is provided in \cref{sec:intro} Introduction.

\noindent\textbf{4)  Model Ensembling} This approach utilizes multi-expert architectures and aggregates the predictions of several specialized models to address LTR challenges. 
\cite{Zhou_2020_CVPR} introduced the Bilateral-Branch Network, a dual-branch architecture. One branch focuses on learning general feature representations, while the other is a re-balanced branch specifically dedicated to learning an effective classifier for tail classes. 
\cite{wang2021longtailed} proposed a multi-branch network, Routing Diverse Distribution-Aware Experts, to learn diverse classifiers in parallel.
\cite{NEURIPS2022_dc6319dd} trained several experts with varied skills and aggregated their outputs during inference, resulting in better performance.
\section{Method}
\label{me}
\subsection{Preliminaries}
\label{pre}
We consider a $K$-class classification task. The aim of LTR is to fit a DNN $F(\cdot; \Theta)$ 
on a long-tailed dataset $\mathcal{B} = \{(x_i, y_i)\}_{i=1}^N$, where the sample $x_i$ is labeled as $y_i \in \{1,\dots,K\}$. 
\(\Theta\)  represents all the parameters of the network. \(N\) is the total number/count of samples in the dataset. 
For class $k$, $\mathcal{B}_k$ is the set of all its examples in $\mathcal{B}$, and \(n_{k}\) is the corresponding sample count. The imbalance factor (IF), defined as $\frac{Max(n_k)}{Min(n_k)}$, is used to quantify the degree of imbalance. 
For conciseness, we denote $f_k(\cdot)$ as the logit for class $k$, and its parameters $\theta_k$ include the weight parameter $w_k$ and the bias parameter $b_k$. 
As such, the logit for class \(k\) can be computed from the extracted feature \(\mathbf{x}_{i}\) (the output of the penultimate layer) as:
\begin{equation}
\begin{aligned}
f_{k}(\mathbf{x}_{i})=&w_{k}^{T} \mathbf{x}_{i}+b_{k}=||w_{k}||_2\cdot ||\mathbf{x}_{i}||_2
\cdot \cos \alpha_k + b_{k},
\end{aligned}
\label{eq:1}
\end{equation}
where \(\alpha _{k}\) denotes the angle between the weight vector \(w_{k}\) and the feature vector \(\mathbf{x}_{i}\). 
Applying the softmax function yields the posterior probability:
\begin{equation}
p(y_i | {\mathbf{x}_{i}})=\frac{e^{f_{y_i}(\mathbf{x}_{i})}}{\sum_{k=1}^{K} e^{f_{k}(\mathbf{x}_{i})}}.
\label{eq:2}
\end{equation}
When Cross-Entropy (CE) loss is used for training, the loss function \(L\) for a single sample is defined as:
\begin{equation}
L=-\sum^{K}_{k=1}\mathbb{I}(y_{i}=k)\log(\frac{e^{f_{k}(\mathbf{x}_{i})}}{\sum_{j=1}^{K} e^{f_{j}(\mathbf{x}_{i})}}),
\end{equation}
where the indicator function \(\mathbb{I}(\cdot )\) equals 1 if \(y_{i}=k\), and 0 otherwise. The network parameters \(\Theta \) are optimized by minimizing the average loss over the entire training set \(\mathcal{B}\): 
\begin{equation}
\label{eq:objective}
\Theta^\ast \;=\; \arg\min_{\Theta}\; F(\Theta;\mathcal{B})
\;\equiv\;
\frac{1}{N}\sum_{i=1}^N L\big(f(\mathbf{x}_i;\Theta),\,y_i\big).
\end{equation}

  


\subsection{Overfitting versus Underfitting}
\label{ou}

In LTR, some works have found that rare classes have significantly smaller weight norms than head classes \cite{Alshammari_2022_CVPR,10105458,Dang_Yang_Dong_Li_Shi_2024}. While this observation may indicate representation underfitting among rare classes, prior work offers conflicting explanations. A number of papers attribute poor results to underfitting in rare classes \cite{Alshammari_2022_CVPR,10105458,Dang_Yang_Dong_Li_Shi_2024,10105459,LI20181}, but several others report overfitting in those same classes \cite{Ye_2021_ICCV,7860349,DBLP:journals/corr/abs-2001-01385,9081988}.
Evidently, the optimal strategy for rescaling weight norms hinges directly on resolving this fundamental debate. In this section, we provide insight into the class-conditional distribution to further explore and clarify this issue. 

Assuming \(h\) and \(t\) denote a head class and a rare class, respectively, where \(n_{h}\gg n_{t}\). Their shared decision boundary \(l\) is implicitly defined by the condition \(f_{h}(\mathbf{x}_{i})=f_{t}(\mathbf{x}_{i})\). Substituting the logit expression from \cref{eq:1}, this boundary can be represented as the locus of points \(\mathbf{x}_{i}\) satisfying:
\begin{equation}
\begin{aligned}
l: &f_h(\mathbf{x}_i) = f_t(\mathbf{x}_i)\\
&(||w_h||_2 \cdot \cos \alpha_h - ||w_t||_2 \cdot \cos \alpha_t) \cdot ||\mathbf{x}_i||_2
\\& + (b_h - b_t) = 0.
\end{aligned}
\end{equation}
The cumulative gradient magnitudes received by each classifier weight \(w_{k}\) and bias \(b_{k}\) are approximately proportional to the class frequency \(n_{k}\), which is the update frequency \cite{10.5555/1953048.2021068}. This differential updating results in a general trend where \(\|w_{k}\|_{2}\) and \(b_{k}\) scale proportionally to \(n_{k}\). Consequently, the decision boundary \(l\) is pushed closer to the rare class \(t\). In other words, as the frequency of a class increases during training, its corresponding weight norm and bias generally increase. 
By applying the law of total probability, the posterior probability from \cref{eq:2} can be rewritten in terms of class-conditional distributions and class priors
\begin{equation}
\begin{aligned}
p(y_i|\mathbf{x}_{i})=\frac{e^{f_{y_i}(\mathbf{x}_{i})}}{\sum_{k=1}^{K} e^{f_{k}(\mathbf{x}_{i})}} 
=\frac{\mathcal{D}^{y_i}_{\mathrm{x}}\left(\mathbf{x}_{i}\right) p(y_i)}{\sum_{k=1}^{K} \mathcal{D}^{k}_{\mathrm{x}}\left(\mathbf{x}_{i}\right) p(k)},
\end{aligned}
\label{eq:6}
\end{equation}
where $\mathcal{D}^{k}_{\mathrm{x}}\left(\mathbf{x}_{i} \right)$ represents the class-conditional distribution of features for class \(k\), and $p(k)$ is the prior probability of class $k$.
Since the prior probability is fixed by the dataset and remains unchanged during the training process, the term $e^{f_{k}(\mathbf{x}_{i})}$ can be interpreted as the unnormalized evidence for the class-conditional distribution $\mathcal{D}^{k}_{\mathrm{x}}\left(\mathbf{x}_{i}\right)$. 
Consequently, the posterior probability \(p(k|\mathbf{x}_{i})\) is proportional to the class-conditional distribution, which in turn is proportional to the exponential of the logit: 
\begin{equation}
\begin{aligned}
 p(k|\mathbf{x}_{i}) \propto \mathcal{D}^{k}_{\mathrm{x}}\left(\mathbf{x}_{i} \right) \propto e^{f_{k}(\mathbf{x}_{i})} = e^{w_{k}^{T} \mathbf{x}_{i}+b_{k}}.
\label{eq:7}
\end{aligned}
\end{equation}

Let \(Z_{k}\) be the proportionality constant that relates the class-conditional distribution \(\mathcal{D}_{\mathrm{x}}^{k}\left(\mathbf{x}_{i}\right)\) to its unnormalized evidence \(e^{f_{k}(\mathbf{x}_{i})}\). 
According to \cref{eq:1}, the value of \(e^{f_{k}(\mathbf{x}_{i})}\) increases as the cosine similarity between the feature vector \(\mathbf{x}_{i}\) and the weight vector \(w_{k}\) increases (i.e., as the angle \(\alpha _{k}\) decreases). 
If \(w_{k}\) is reduced by a factor \(\alpha \) where \(0<\alpha <1\), the new weight vector \(w_{k}^{\prime }\) and the new logit \(f_{k}^{\prime }(\mathbf{x}_{i})\) are: 
\begin{equation}
\begin{aligned}
w_k'=\alpha \cdot w_k,
\label{eq:8}
\end{aligned}
\end{equation}
\begin{equation}
\begin{aligned}
f'_k(\mathbf{x}_i)=\alpha \cdot w_k^\top \mathbf{x}_i + b_k,
\label{eq:9}
\end{aligned}
\end{equation}
Combining these with the proportionality from \cref{eq:7}, the new class-conditional distribution \(\mathcal{D}_{\mathrm{x}}^{\prime k}(\mathbf{x}_{i})\) is expressed as:
\begin{equation}
\label{eq:10}
\mathcal{D'}^{k}_{\mathrm{x}}\left(\mathbf{x}_{i} \right) \propto e^{f'_k(x)} = e^{(1-\alpha)b_k} (Z_k)^{\alpha} [D_\mathbf{x}^k(\mathbf{x}_i)]^{\alpha},
\end{equation}
After normalization over $\mathbf{x}$, the implicit class-conditional distribution is therefore
\begin{equation}\label{eq:11}
\mathcal{D'}^{k}_{\mathrm{x}}\left(\mathbf{x}_{i} \right) = \frac{[\mathcal{D}^{k}_{\mathrm{x}}\left(\mathbf{x}_{i} \right)]^{\alpha}}{\int [D_\mathrm{x}^k(\mathbf{x})]^{\alpha} d\mathbf{x}}.
\end{equation}
As demonstrated in \cref{eq:11}, because \(0<\alpha <1\), reducing the per-class weight norm \(||w_{k}||_{2}\) (which corresponds to rarer classes) is equivalent to performing a ``power-law contraction" on the original distribution \(\mathcal{D}_{\mathrm{x}}^{k}\left(\mathbf{x}_{i}\right)\). 
The smaller $||w_k||_2$ is, the smoother and more diffuse the resulting class-conditional distribution $\mathcal{D'}^{k}_{\mathrm{x}}\left(\mathbf{x}_{i}\right)$ becomes. In the extreme case, if \(\alpha \) is close to zero, the harvested distribution approaches a uniform distribution. This indicates that little knowledge has been obtained from the samples of the rare class. The detailed mathematical derivation can be found in Appendix A.
 According to the analysis above, instead of learning overfitting features in rare classes, the model learns underfitting features, leading to poor performance in LTR.
These analyses provide additional evidence for the norm rescaling strategy from a probabilistic perspective.

\subsection{Self-Adaptive Monotonic Normalization}
Having established that poor performance in LTR stems from the representation underfitting of rare classes—which provides a strong theoretical basis for norm rescaling—our objective is to address the limitations of current rescaling techniques. In general, many of these methods aim to compensate for representation underfitting by explicitly enlarging the weight norms of the rare classes, thereby expanding their decision boundaries. 
However, most norm rescaling approaches \cite{09217,Zhong_2021_CVPR,Alshammari_2022_CVPR,10105458,Dang_Yang_Dong_Li_Shi_2024,Du_2023_CVPR}. achieve this through parameter regularization, which inevitably introduces hyperparameters that require careful tuning. As noted previously, LTR performance is highly sensitive to these specific hyperparameter configurations. 
Consequently, there is an urgent need for a robust method that can deliver outstanding performance without relying on complex hyperparameter selection. Here we propose a simple yet effective approach: Self-Adaptive Monotonic Normalization (SAMN).

Specifically, we suggest that the per-class weight norms should generally increase from head classes to rare classes to effectively compensate for this imbalance. The critical challenge is maintaining this required increase throughout training without introducing any additional hyperparameters. We achieve this by proposing a self-adaptive mechanism that enforces this ordering constraint directly on the learnable weights.
To achieve this, we decouple the classifier weights into direction and magnitude, enforcing monotonicity on the applied magnitude. As such, the SAMN method computes the logit \(f_{k}(\mathbf{x}_{i})\) using modified parameters \(\hat{w}_{k}\) and \(\hat{b}_{k}\): 
\begin{equation}
\begin{aligned}
f_{k}(\mathbf{x}_{i})=\hat{w}_{k}^{T} \mathbf{x}_{i}+\hat{b}_{k},
\end{aligned}
\label{eq:12}
\end{equation}
\begin{equation}
\begin{aligned}
\hat{w}_{k} = \frac{w_k}{||w_k||_2} \cdot e^{s^w_k},
\end{aligned}
\label{eq:13}
\end{equation}
\begin{equation}
\begin{aligned}
\hat{b}_{k} = e^{s^b_k} + b_k,
\end{aligned}
\label{eq:14}
\end{equation}
where $s^w_k$ and $s^b_k$ are learnable class-wise scalar parameters for the weight and bias of class \(k\), respectively. 
These scalars are constrained to be monotonic based on a predefined order metric, such as class frequency. 
 To ensure that the final norm is positive and greater than one, and to amplify the gradients of rare classes during the second stage, we use the exponential function to obtain the final scaling factor.

 \begin{algorithm}[!t]
\small
\caption{\textbf{PAVA Process in SAMN}}
\label{alg:1}
\KwIn{Order Metric Sequence $\mathbf{A} = [a_1, \dots, a_K]$, Raw Learnable Parameters $\mathbf{R} = [r_1, \dots, r_K]$.}
\KwOut{Scaling Factor Sequence $\mathbf{S} = [s_1,...,s_K]$.}

\BlankLine
\textbf{Step 1: Sort Classes by the Order Metric}\\
$(\textit{sorted\_A}, \textit{perm}) \gets \text{sort}(\mathbf{A})$\\
$\textit{inv\_perm} \gets \text{inverse\_permutation}(\textit{perm})$\\
$\textit{reordered\_R} \gets \mathbf{R}[\textit{perm}]$

\BlankLine
\textbf{Step 2: PAVA Projection (Isotonic Regression)}\\
Initialize empty lists: $\textit{sums} \gets [\,], \textit{lengths} \gets [\,]$\\
\For{$i = 1$ \textbf{to} $K$}{
    $\textit{m} \gets \textit{reordered\_R}[i],\ \textit{l} \gets 1$\\
    Append $\textit{m}$ to \textit{sums}, and $\textit{l}$ to \textit{lengths}\\
    \While{$|\text{sums}|
 \ge 2$ \textbf{and} 
        $\frac{\text{sums}[-2]}{\text{lengths}[-2]} > \frac{\text{sums}[-1]}{\text{lengths}[-1]}$}{
        Merge last two blocks:\\
        $\textit{sums}[-2] \gets \textit{sums}[-2] + \textit{sums}[-1]$\\
        $\textit{lengths}[-2] \gets \textit{lengths}[-2] + \textit{lengths}[-1]$\\
        Remove the last entries from both lists
    }
}
$proj\_sorted \gets []\;$ \\
\For{$j = 1$ \KwTo $\text{length}(sums)$}{ 
    $block\_mean \gets \frac{sums[j]}{lengths[j]}\;$ \\
    $block\_length \gets lengths[j]\;$ \\
    \For{$z = 1$ \KwTo $block\_length$}{
        Append $block\_mean$ to $proj\_sorted\;$
    }
}
\BlankLine
\textbf{Step 3: Map to Positive Values and Revert Order}\\
$\mathbf{U} \gets \text{softplus}(\textit{proj\_sorted})$\\
$\mathbf{S} \gets \mathbf{U}[\textit{inv\_perm}]$

\Return $\mathbf{S}$
\end{algorithm}

The learnable scalars \(s_{k}^{w}\) and \(s_{k}^{b}\) are central to SAMN.
As highlighted previously, successful compensation for rare classes requires ensuring that their effective weight norms remain greater than those of abundant head classes throughout the adaptive learning process.  Here we employ the Pool Adjacent Violators Algorithm (PAVA) \cite{ff03617f-3b30-3cb3-a0fc-02979e761de5} to enforce this monotonic constraint. 
PAVA is a \textit{hyperparameter-free} and efficient method used to solve the isotonic regression problem. Specifically, given an arbitrary set of observations $\mathbf{R} = [ r_1, r_2, \dots, r_n ]$ (which, in our case, are the values of the raw learnable scalars), the algorithm finds a non-decreasing sequence $\mathbf{S} = [s_1 \le s_2 \le \dots \le s_n ]$ that minimizes the sum of squared errors:
\begin{equation}
\min_{s_1 \le s_2 \le \dots \le s_n} 
\sum_{i=1}^{n} (r_i - s_i)^2.
\end{equation}
By applying PAVA, we ensure that optimized scalars \(s_{i}\) inherently satisfy the desired monotonic order based on an order metric (such as class frequency), eliminating the need for sensitive complex hyperparameter tuning.
In the SAMN procedure, we first obtain an initial set of learnable norms and reorder them according to the sorted order metric sequence \(\mathbf{A}\) (e.g., in ascending order of class frequency). Next, PAVA is applied to this reordered sequence to enforce strict monotonicity, solving the isotonic regression problem. Subsequently, the order of this newly monotonic sequence is reverted to match the original class indices, yielding the final scaling factor sequences \(\mathbf{S}^{w}\) and \(\mathbf{S}^{b}\) (comprising \(s_{k}^{w}\) and \(s_{k}^{b}\) for all classes \(k\)). The process is detailed in \cref{alg:1}.
\begin{figure}[!t]
  \centering 
  \includegraphics[width=0.82\columnwidth]{rebuttal_pict/robust_complete_legend2.png}
  \begin{subfigure}{0.42\linewidth}
    \includegraphics[width=\linewidth]{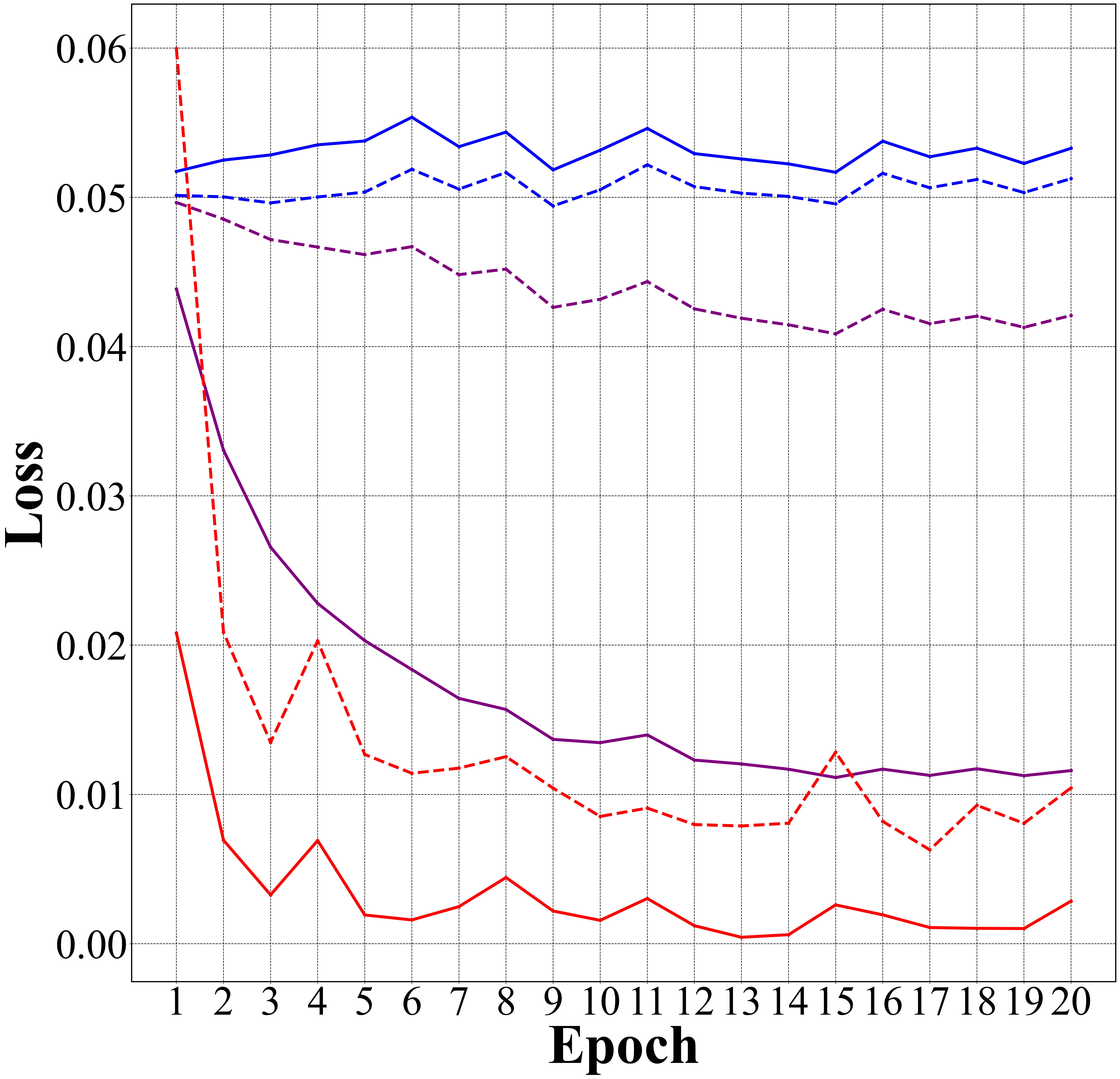}
    \caption{CIFAR100}
    \label{fig:1a_add}
  \end{subfigure}
  \hspace{0.3cm}
  \begin{subfigure}{0.42\linewidth}
    \includegraphics[width=\linewidth]{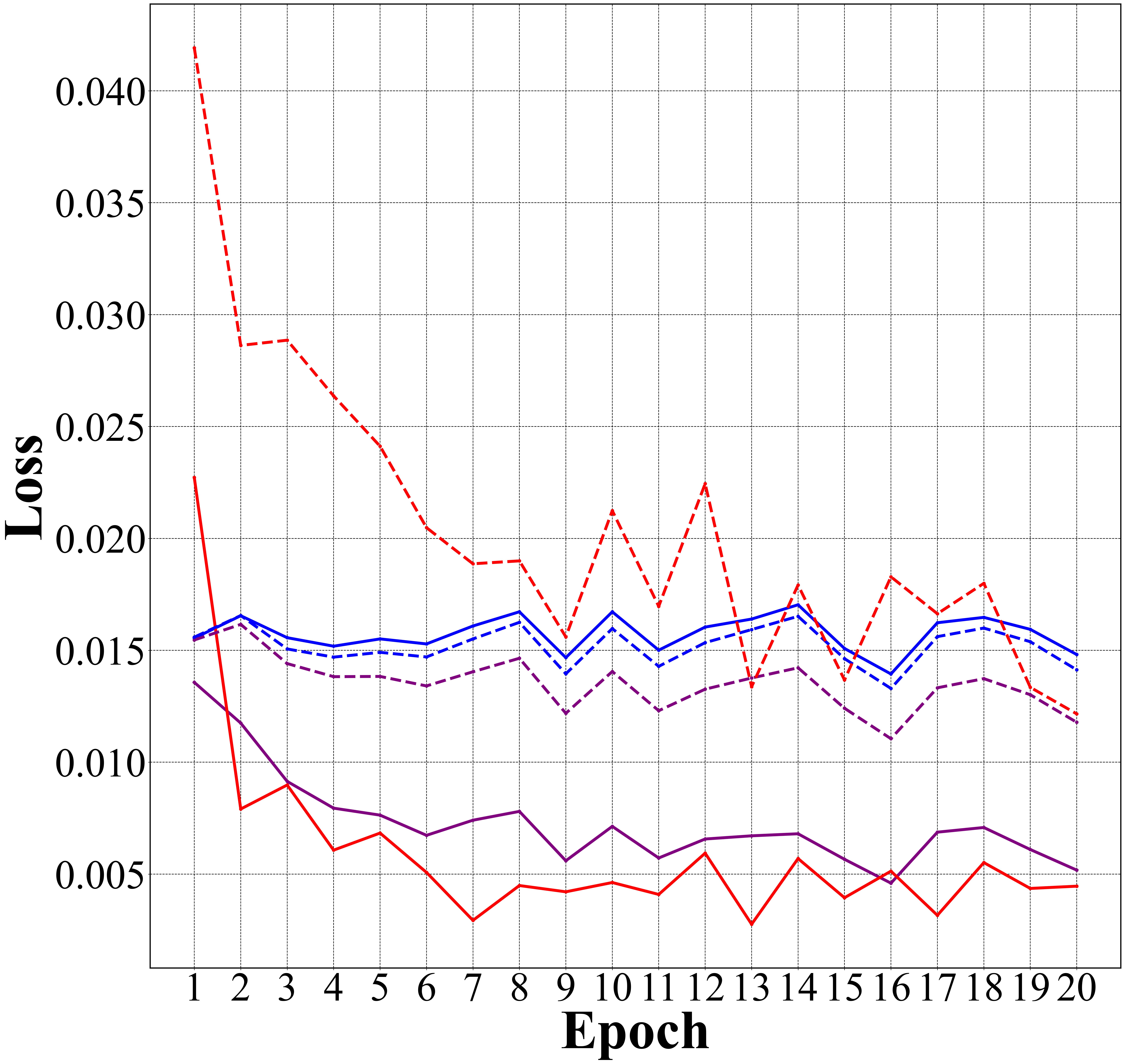}
    \caption{CIFAR10}
    \label{fig:1b_add}
  \end{subfigure}
\caption{Loss of various methods during retraining with IF = 100. Solid and dashed lines represent the learning rate of 0.01 and 0.0005, respectively.}
\label{fig:1add}
\end{figure}

In this work, we provide two effective order metrics for sorting the classes before applying PAVA: 1) Class Frequency: Our prior analysis indicated that weight norms typically increase with class count during standard training. However, because our goal is to compensate for underfitting by ensuring rare classes have larger final norms, we use the inverted class frequency (i.e., \(\frac{1}{n_{k}}\)) as the order metric to align with PAVA's non-decreasing output constraint. 2) First-Stage Weight Norms: A more adaptive approach uses the actual inverted per-class weight norms obtained after the representation learning stage. This leverages the model's organically learned imbalance status as the sorting basis. For both situations, the raw learnable parameters \(\mathbf{R}\) in \cref{alg:1} are initialized corresponding to the chosen metric (either the inverted class frequency values or the inverted first-stage weight norms). It should be noted that the selection for the order metric can be regarded as a categorical hyperparameter. However, compared to the continuous hyperparameters in previous works, this categorical parameter in SAMN is significantly easier to set. Moreover, all available options for this hyperparameter demonstrate performance approaching or even surpassing most state-of-the-art methods (see \cref{tab:1,tab:3}). Therefore, SAMN is still considerably more hyperparameter-friendly than previous methods.

Notably, PAVA is introduced in SAMN for the following two reasons: 1) Flexibility and Local Malleability -- PAVA ensures that the changes of weight norms across classes are relatively flexible. Consider a more rigid approach, such as using a positive learnable parameter \(p_{k}\) to enforce monotonicity recursively (\(s_{k+1}=s_{k}+p_{k+1}^{2}\)). In that scenario, \(s_{k+1}\) encompasses the entirety of all preceding terms (\(s_{1},\dots, s_{k}\)), leading to a highly entangled and rigid scaling sequence \(\mathbf{S}\). 
In contrast, PAVA's ``pooling" operation acts locally; a given \(s_{k}\) only forms relationships with its adjacent pooled classes. This structure ensures that each individual weight norm remains as malleably learnable as possible while still adhering to the monotonic constraints.
2) Leveraging Prior Knowledge -- The sequence produced by PAVA is guaranteed to be the isotonic projection that is most similar to the original input sequence in terms of Euclidean distance. Since our initial raw learnable parameters \(\mathbf{R}\) are initialized using valuable prior knowledge (either class frequency or first-stage weight norms), employing PAVA allows us to effectively leverage this crucial prior information while enforcing monotonicity. 

Although PAVA is an abrupt manual adjustment to the weight norm, it does not disrupt the convergence as it is a non-expansive operator. As shown in \cref{fig:1add}, we compare the SAMN against CE and Weight Decay (WD). SAMN converges smoothly without significant oscillations, matching the stability of baselines across various learning rates.

\begin{figure*}[!t]
  \centering
  \begin{subfigure}{0.22\linewidth}
     \includegraphics[width=\linewidth]{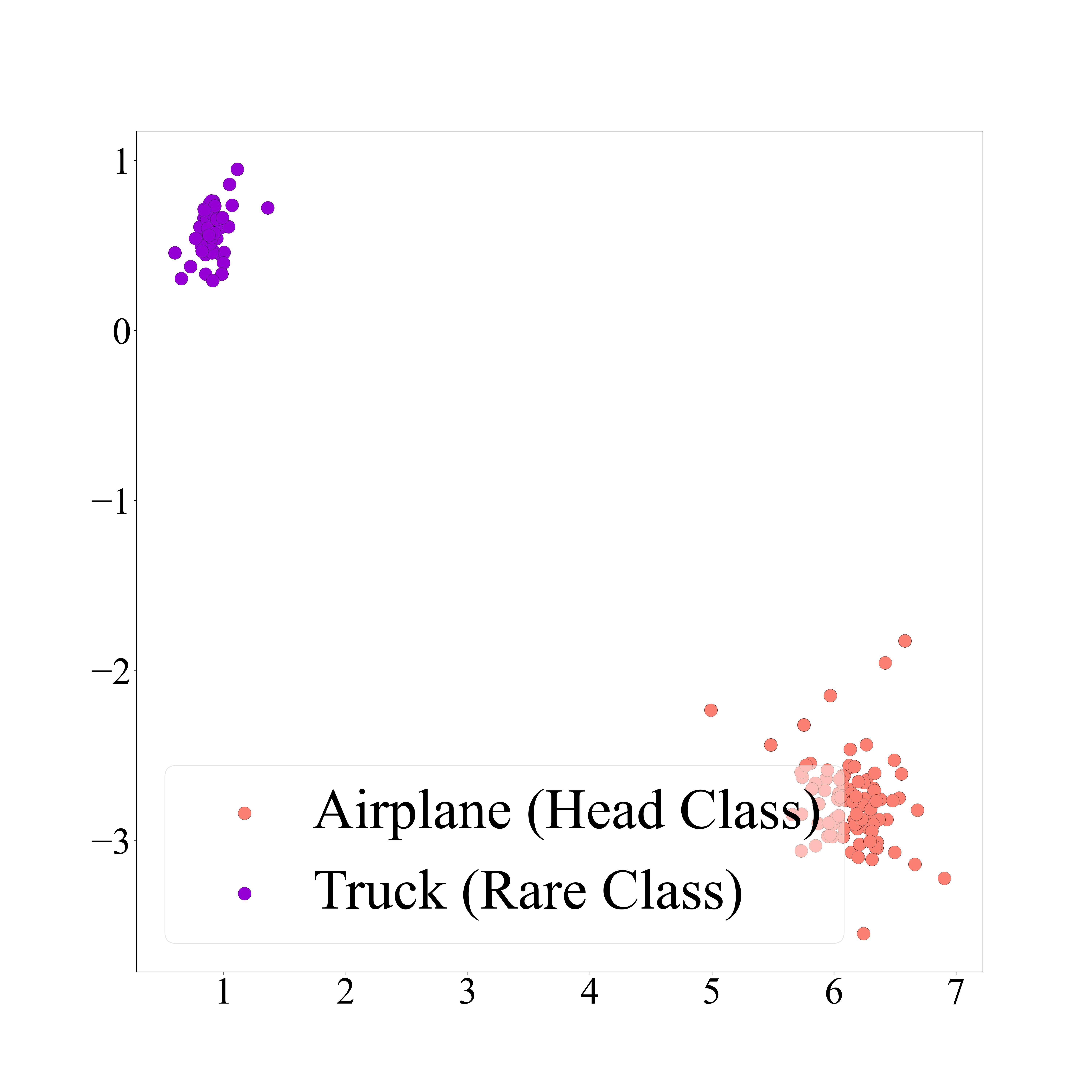}
    \caption{Data distribution in training set}
    \label{fig:4a}
  \end{subfigure}
  \begin{subfigure}{0.22\linewidth}
     \includegraphics[width=\linewidth]{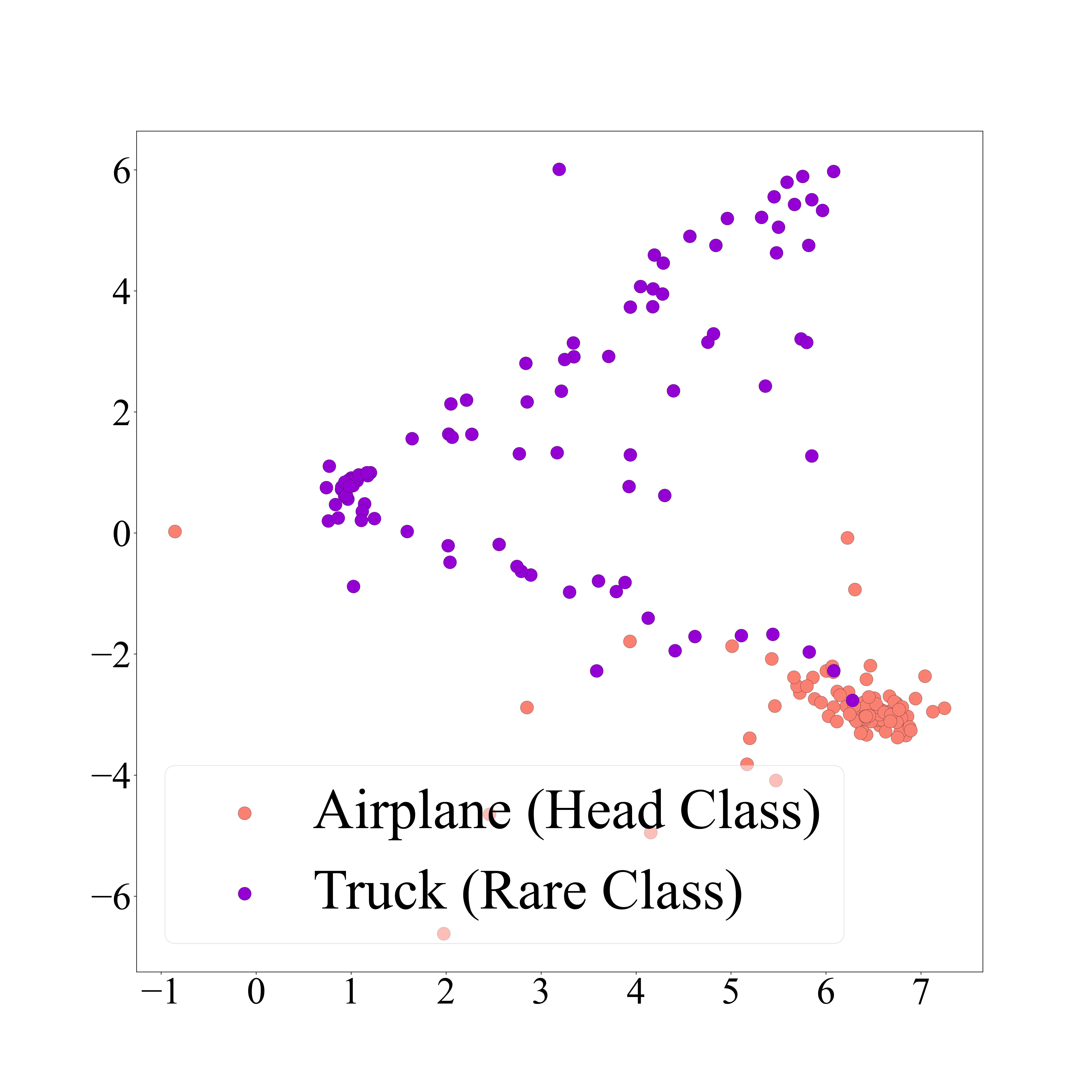}
    \caption{Data distribution in test set}
    \label{fig:4b}
  \end{subfigure}
  \begin{subfigure}{0.22\linewidth}
    \includegraphics[width=\linewidth]{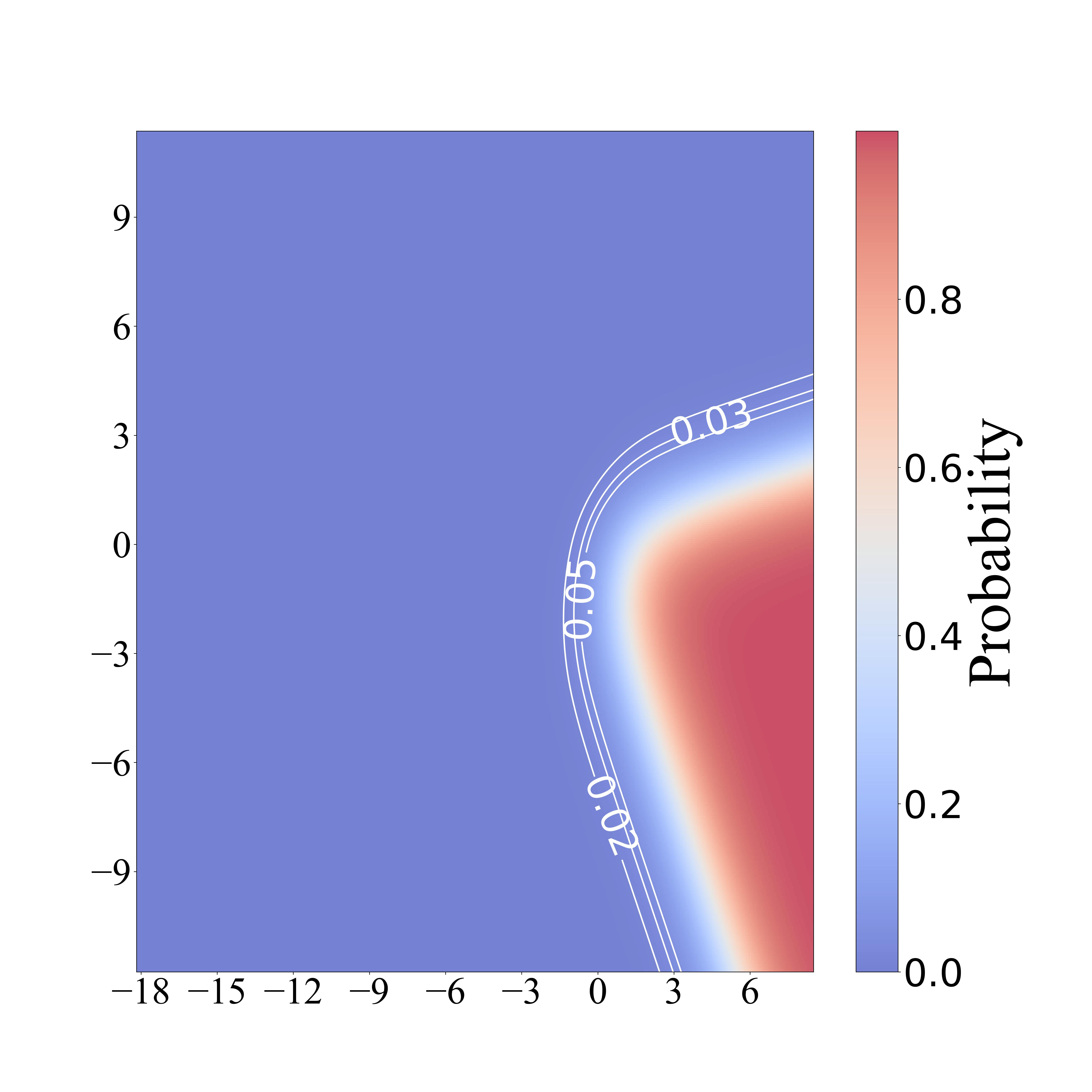}
    \caption{Head class distribution}
    \label{fig:4c}
  \end{subfigure}
  \begin{subfigure}{0.22\linewidth}
    \includegraphics[width=\linewidth]{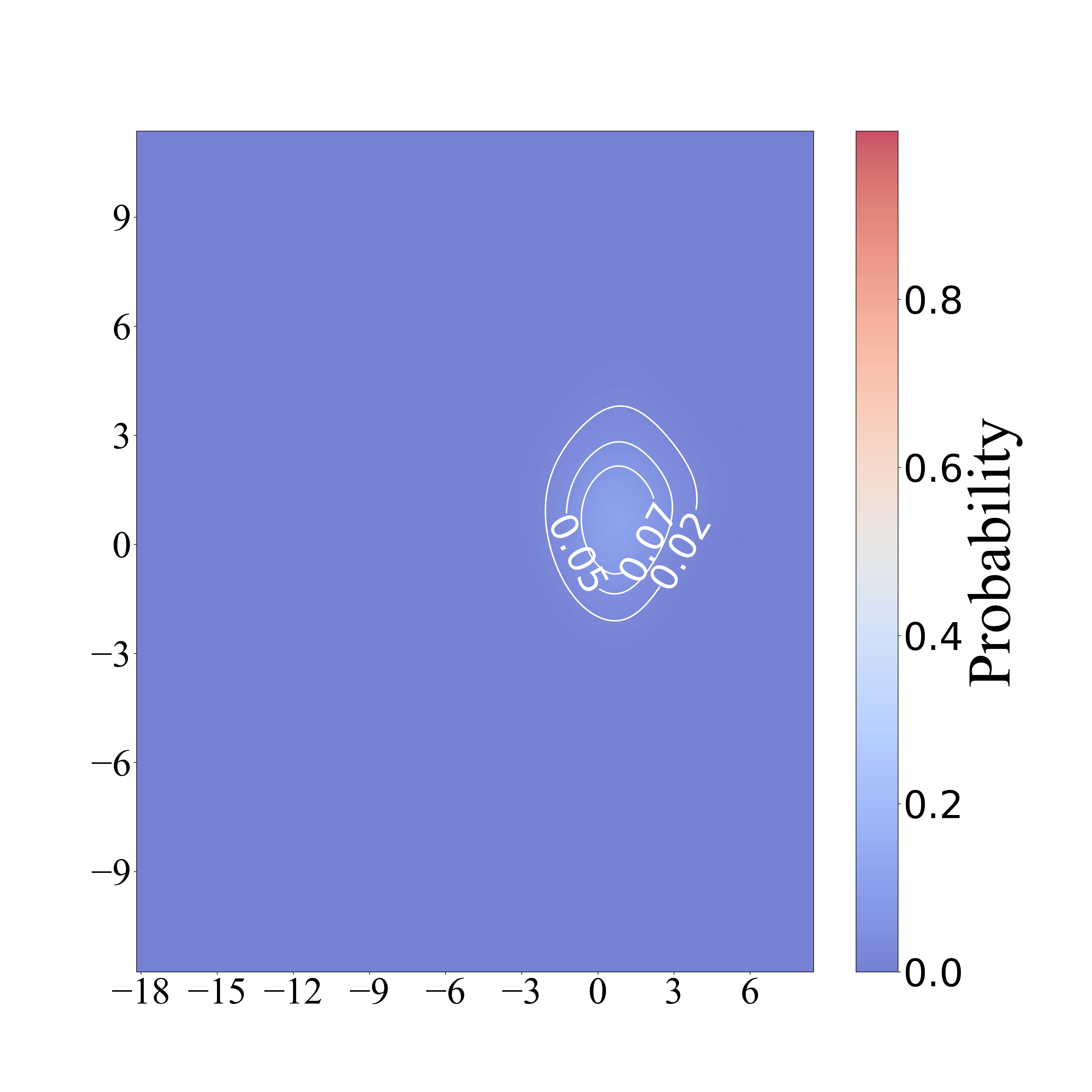}
    \caption{Rare class distribution}
    \label{fig:4d}
  \end{subfigure}
  \caption{Data and class-conditional distributions of the head class and the rare class. In the training set, the head and rare classes are clustered, while the rare class is dispersed in the test set. The head class shows a sharply peaked class-conditional distribution, while that of the rare class is much flatter. These results support our theoretical analysis in \cref{me}. }
  \label{fig:4}
\end{figure*}

\begin{figure}[!t]
  \centering 
    \includegraphics[width=0.9\columnwidth]{rebuttal_pict/robust_complete_legend3.png}
  \begin{subfigure}{0.42\linewidth}
    \includegraphics[width=\linewidth]{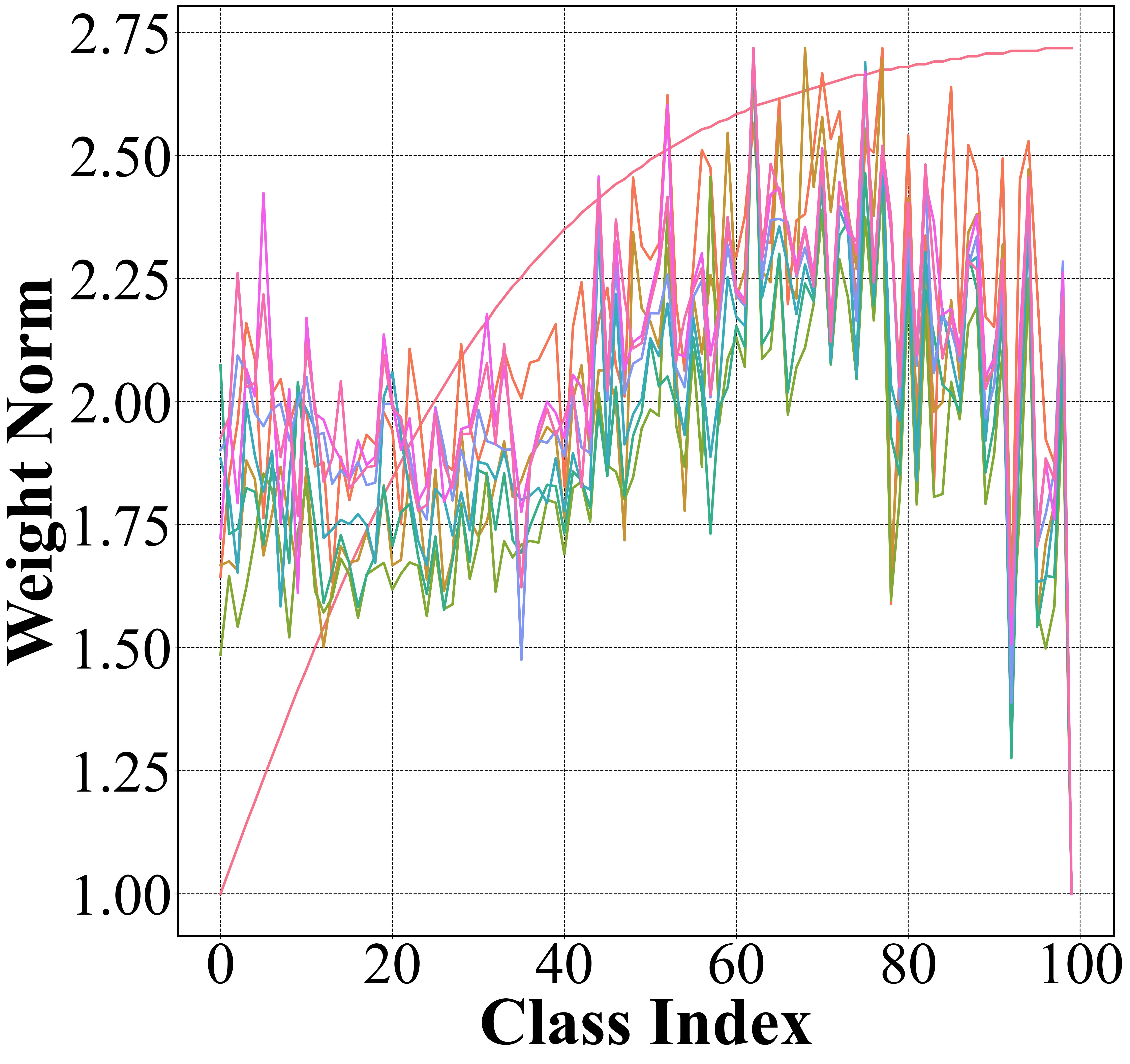}
    \caption{CE}
  \end{subfigure}
  \hspace{0.7cm}
  \begin{subfigure}{0.42\linewidth}
    \includegraphics[width=\linewidth]{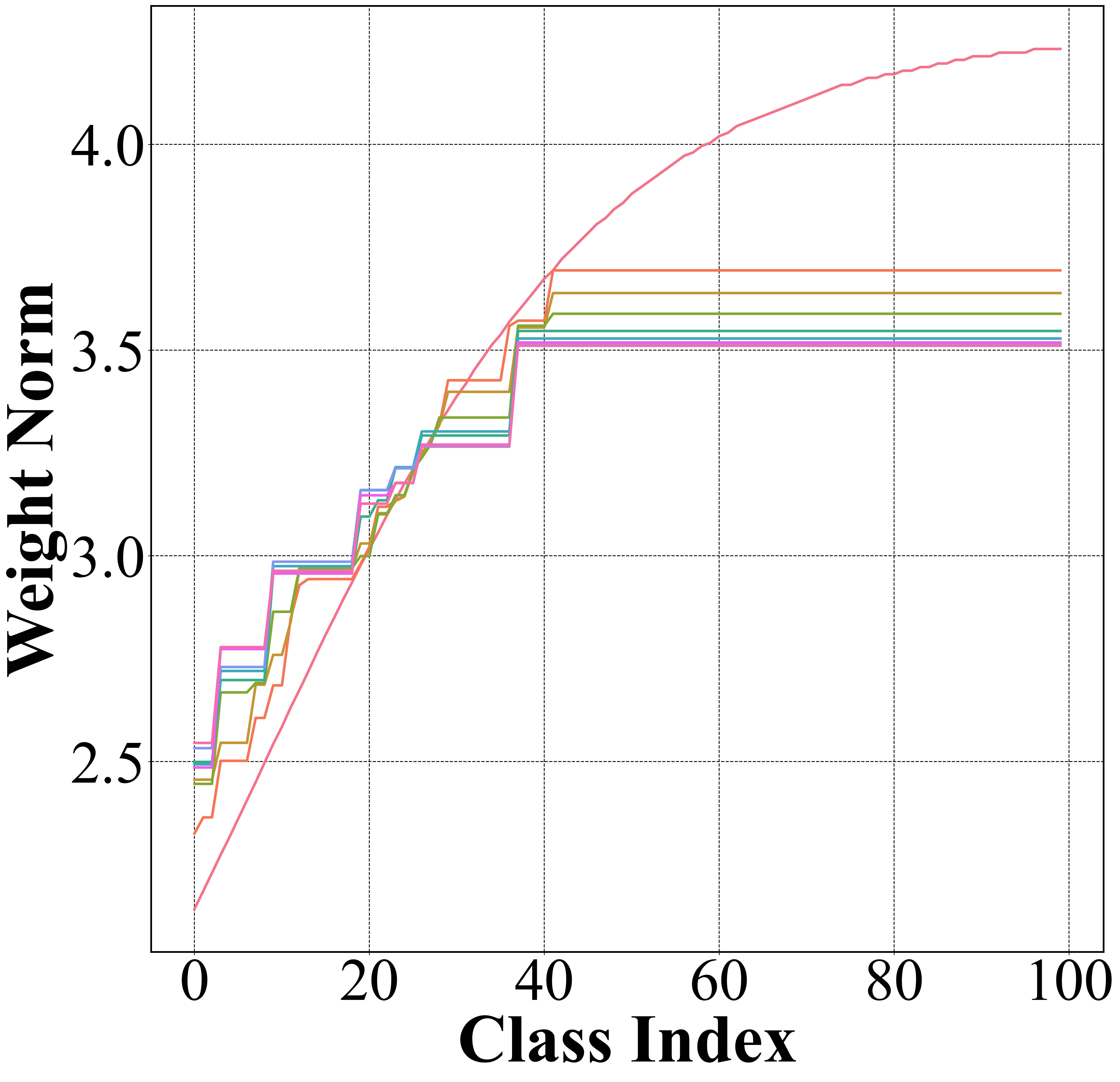}
    \caption{SAMN}
  \end{subfigure}
\caption{Evolution of weight norms on CIFAR100 with IF = 100.}
\label{fig:2add}
\end{figure}

\section{Experiments}
\label{ex}
We carry out a series of experiments to reveal the characteristics of SAMN in LTR.
Firstly, we perform empirical studies to demonstrate the underfitting of rare classes in LTR and the hyperparameter-friendly advantage of our method.
Then, we compare SAMN with the state-of-the-art (SOTA) methods on four long-tailed datasets.
Finally, we conduct ablation studies to discuss the design choices in our approach. The source code will be made publicly available.

\noindent\textbf{Datasets and Evaluation Metrics.} We employ CIFAR10-LT \cite{10.5555/1953048.2021069}, CIFAR100-LT \cite{10.5555/1953048.2021069}, ImageNet-LT \cite{Liu_2019_CVPR}, iNaturalist2018 \cite{Horn_2018_CVPR} to perform the experiments.
Regarding the evaluation metric, we use the accuracy in all datasets.
The detailed description can be found in Appendix B.

\subsection{Empirical Study}
\label{eu}
The study is divided into three parts:
1) Underfitting Demonstration: We analyze sample distributions and class-conditional evidence in the hidden space (\cref{eq:7} and \cref{eq:11}) to empirically demonstrate that poor LTR performance is attributed to the underfitting of rare classes; 2) Retraining Dynamics: We observe the evolution of weight norms during retraining to support the better performance after using our method; and 3) Hyperparameter Sensitivity Analysis: We use SOTA norm rescaling methods as examples to illustrate their sensitivity to hyperparameters, contrasting this with our proposed SAMN method.

\noindent\textbf{Experimental Setup.} For the first part, experiments were conducted on CIFAR10-LT with IF=100.
We used a ResNet32 \cite{He_2016_CVPR} model whose dimension of the penultimate output was two for better visualization.
It was trained for 200 epochs with a batch size of 64 and a WD \cite{NIPS1988_1c9ac015} of 5e-3.
The stochastic gradient descent (SGD) optimizer with momentum 0.9 and the cosine learning rate scheduler \cite{DBLP:journals/corr/LoshchilovH16a} were applied for training, and the initial learning rate was 0.01. For the second part, after training with the same condition mentioned above, we separately retrained the model using CE and SAMN to explore. The learning rate was set to 0.05; other setups are described in \cref{cs}.
For the third part, we used the standard ResNet32 \cite{He_2016_CVPR} to explore.
After training with the same condition mentioned above, we separately retrained the model using WD \cite{NIPS1988_1c9ac015} and Shifted Label-Aware Smoothing (SLAS) \cite{Zhong_2021_CVPR} with various hyperparameter values.
For WD, we set its hyperparameter to \{1e-4, 1\}.
For SLAS, we separately set the hyperparameters $\epsilon_k$ and $\epsilon_1$ in \cite{Zhong_2021_CVPR} to zero and \{1e-4, 1\}.
For better comparison, we also report the result of SAMN, and its setup is described in \cref{cs}.

\begin{table*}[!t]
\centering
\caption{Accuracy (\%) on CIFAR100-LT and CIFAR10-LT datasets.
The imbalance factor is set to 100, 50, and 10. We reproduce the original methods that are plugged by SAMN for fair comparison. WD: Weight Decay. *: reproduced results. The best results are in \textbf{bold}.}
\label{tab:1}
\scalebox{0.728}
{
\begin{tabular}{lccccccc}
\toprule
\multirow{2}{*}{Method} & \multicolumn{3}{c}{CIFAR100-LT} & \multicolumn{3}{c}{CIFAR10-LT}\\
\cmidrule(lr){2-4} \cmidrule(lr){5-7}
 & 100 & 50 & 10 & 100 & 50 & 10 \\
\midrule
CE (with WD in stage one) & 47.4* & 52.3* & 67.2* & 80.0* & 84.3* & 91.9* \\
Focal Loss \cite{Lin_2017_ICCV}  & 38.4 & 44.3 & 55.8 & 70.4 & 76.7 & 86.7  \\
CAM-BS \cite{Zhang_Wei_Zhou_Wu_2021}  & 41.7 & 46.0 & - & 75.4 & 81.4 & -  \\
LDAM-DRW \cite{NEURIPS2019_621461af}   & 42.0 & 46.6 & 58.7 & 77.0 & 81.0 & 88.2  \\
BBN \cite{Zhou_2020_CVPR}  & 42.6 & 47.0 & 59.1 & 79.8 & 82.2 & 88.3 \\
cRT \cite{09217}   & 45.3 & 46.8 & 58.1 & 75.7 & 80.4 & 88.3  \\
DiVE \cite{He_2021_ICCV}  & 45.4 & 51.1 & 62.0 & - & - & -\\
SAM  \cite{NEURIPS2022_8f4d70db} & 45.4 & -  & - & 81.9 & - & -  \\
CSA \cite{NEURIPS2023_eeffa70b}  & 46.6 & 51.9 & 62.6 & 82.5 & 86.0 & 90.8 \\
ADRW \cite{NEURIPS2023_973a0f50}   & 46.4 & - & 61.9 & 83.6 & - & 90.3\\
CMO \cite{Park_2022_CVPR}  & 47.2 & 51.7 & 58.4 & - & - & -  \\
CUDA \cite{ahn2023cuda} & 47.6 & 51.1 & 58.4 & - & -  \\
RIDE (3 experts) \cite{wang2021longtailed}  & 48.0 & - & - & - & - & - \\
DiffuLT \cite{shao2024diffultmakediffusionmodel}  & 51.5 & 56.3 & 63.8 & 84.7 & 86.9 & 90.7  \\
SLAS \cite{Zhong_2021_CVPR}  & 52.7* & 56.6* & 69.7* & 84.2* & 87.2* & 92.7* \\
IWB \cite{Dang_Yang_Dong_Li_Shi_2024} & 53.3 
 & 56.3 & - & - & - & -  \\
WD + MaxNorm \cite{Alshammari_2022_CVPR}
 & 53.4 & 57.7 & 68.7 & - & - & -  \\
GLMC \cite{Du_2023_CVPR} & 56.8* & 62.2* & 72.8* & 88.2* & 90.8* & 95.1* \\
\midrule
\rowcolor{gray!20} CE + SAMN (Ours)   & 54.0 ($\uparrow$6.6) & 58.3 ($\uparrow$6.0) & 70.1 ($\uparrow$2.9) & 84.1 ($\uparrow$4.1) & 87.6 ($\uparrow$3.3) & 92.8 ($\uparrow$0.9)\\ \cdashline{1-7}
\rowcolor{gray!20} SLAS + SAMN (Ours)   & 54.1 ($\uparrow$1.4)  & 58.4 ($\uparrow$1.8)  & 70.1 ($\uparrow$0.4)  & 85.5 ($\uparrow$1.3) & 88.5  ($\uparrow$1.3)  & 92.8 ($\uparrow$0.1)    \\
\rowcolor{gray!20}GLMC + SAMN (Ours)   & \textbf{57.7} ($\uparrow$1.1)  & \textbf{64.0} ($\uparrow$1.8) & \textbf{74.1} ($\uparrow$1.3) & \textbf{88.5} ($\uparrow$0.3) & \textbf{91.3} ($\uparrow$0.5) & \textbf{95.2} ($\uparrow$0.1)   \\

\bottomrule
\end{tabular}
}
\end{table*}

\begin{figure}[!t]
  \centering
  \begin{subfigure}{0.465\linewidth}
    \includegraphics[width=\linewidth]{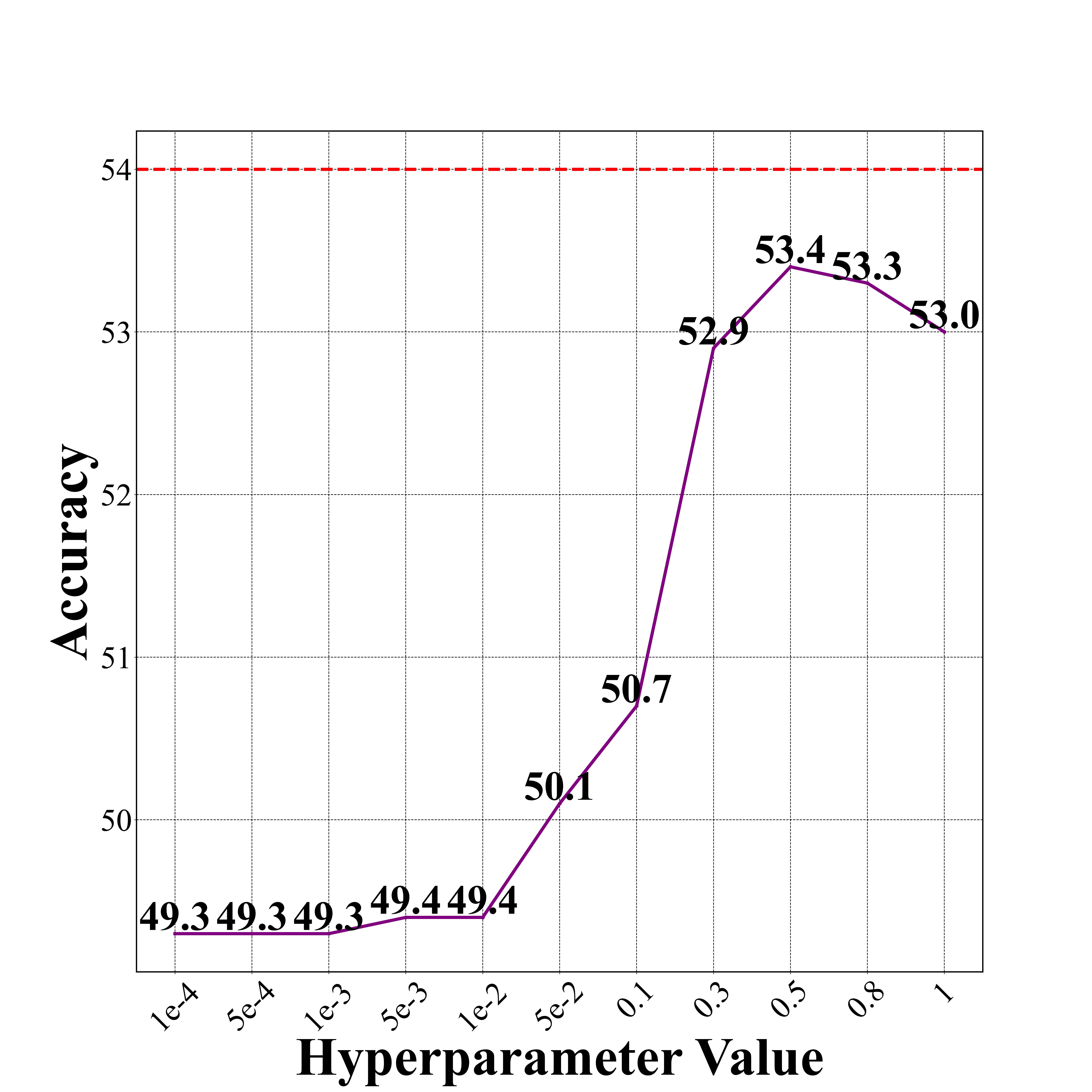}
    \caption{Weight Decay}
    \label{fig:5a}
  \end{subfigure}
    \hspace{0.3cm}
  \begin{subfigure}{0.465\linewidth}
    \includegraphics[width=\linewidth]{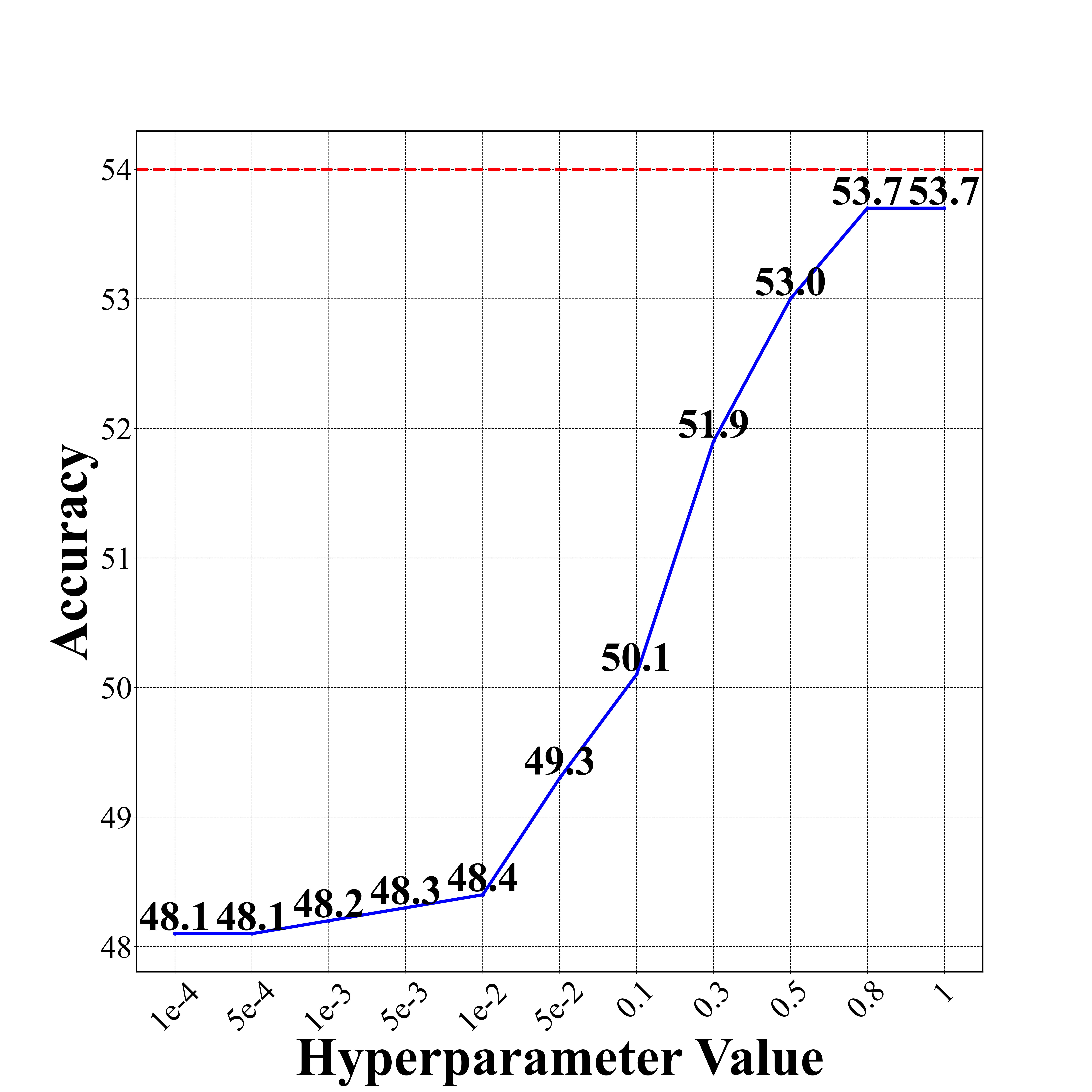}
    \caption{SLAS}
    \label{fig:5b}
  \end{subfigure}
\caption{Accuracy (\%) curves illustrate the performance sensitivity of Weight Decay and SLAS to various hyperparameters with IF = 100. The red dashed lines show the superior and stable performance of our hyperparameter-friendly SAMN, highlighting its robustness.}
  \label{fig:5}
\end{figure}

\noindent\textbf{Results.} As shown in \cref{fig:4a,fig:4b}, in the training set, both samples from the head class \textit{Airplane} and those from the rare class \textit{Truck} are clustered.
However, in the test set, although the samples from the head class \textit{Airplane} are still clustered tightly, those from the rare class \textit{Truck} are scattered. Moreover, as shown in \cref{fig:4c,fig:4d}, the class-conditional distribution of \textit{Airplane} is sharp.
The entire hidden space exhibits distinct regions with high-probability distributions (the red area) and low-probability distributions (the blue area).
Conversely, the class-conditional distribution of the rare class \textit{Truck} is much flatter than that of the \textit{Airplane}.
These results indicate that the head class can be identified by the learned class-conditional distribution, while the rare class cannot; this supports our theoretical analysis in \cref{me} and demonstrates that representation underfitting occurs in the rare class.

A topic requiring further discussion is why rare classes form distinct clusters in the training set (\cref{fig:4a}) despite having diffuse implicit distributions (\cref{fig:4d})? We argue this happens because concentrated head classes (\cref{ou}) leave spatial gaps in the feature space. Rare classes, having flatter distributions, naturally dominate these gap areas by default. Thus, these are ``useless clusters" formed by loss-induced spatial compression (pushing samples away from dominant classes) rather than genuine representation learning, explaining the poor LTR performance.

In terms of retraining dynamics, \cref{fig:2add} visualizes the evolution of weight norms. As can be seen, under CE, although weight norms are initialized by class frequency, they quickly invert, where $||w_{rare}||$ is generally smaller than $||w_{head}||$. SAMN prevents this inversion. Although rare class norms shrink, $||w_{rare}|| \ge ||w_{head}||$ is maintained, leading to superior performance. Regarding robustness, \cref{fig:5} illustrates that WD and SLAS methods demonstrate significant sensitivity to their hyperparameters. A change in the WD hyperparameter from \(1\text{e-}4\) to 1 results in a 4.1\% accuracy change, and SLAS shows a 5.6\% accuracy change. In comparison,  SAMN is hyperparameter-friendly and exhibits robust superior performance (see the red dashed lines in \cref{fig:5}).
\subsection{Comparison with other SOTA methods}
\label{cs}
Following the essential characteristics discussed in our empirical study, we compare the performance of SAMN with SOTA methods on benchmark datasets.

\noindent\textbf{Experimental Setup.} We evaluate SAMN on the benchmark datasets CIFAR10-LT and CIFAR100-LT, and further demonstrate its performance on the large-scale, real-world datasets ImageNet-LT and iNaturalist2018. In CIFAR10-LT and CIFAR100-LT, we adopt ResNet32 \cite{He_2016_CVPR} as the backbone.
In addition to the Cross Entropy (CE) method, we plug SAMN into a one-stage method of Global and Local Mixture Consistency (GLMC) \cite{Du_2023_CVPR} and a two-stage method of Shifted Label-Aware Smoothing (SLAS) \cite{Zhong_2021_CVPR} to observe its generalization capability.
In particular, we first reproduced them with the same training environment for a fair comparison.
After obtaining the models with these methods, we further applied SAMN to the models and retrained them for 20 epochs. 
In ImageNet-LT and iNaturalist2018, we used ResNeXt50 \cite{Xie_2017_CVPR} for training. 
Here, we directly selected GLMC to train the model in the first stage and employed SAMN to improve it in the second stage.
Detailed experimental settings can be found in Appendix C.

\noindent\textbf{Results.} As can be seen in \cref{tab:1}, after being integrated with SAMN, three methods tested achieve better performance.
In CIFAR10-LT, the largest gap is 4.1\%, appearing when SAMN is used in CE with IF=100.
In CIFAR100-LT, the largest gap is 6.6\%, appearing when SAMN is used in CE with IF=100.
The GLMC with SAMN achieves the best results in all tested situations.
In addition, the results on ImageNet-LT and iNaturalist2018 are shown in \cref{tab:2}.
SAMN generally achieves the best performance compared to other SOTA methods.
\begin{table}[!t]
\centering
\caption{Accuracy (\%) on ImageNet-LT and iNaturalist2018 datasets. Best results are in \textbf{bold}. }
\label{tab:2}
\scalebox{0.65}
{
\begin{tabular}{ccccccccc}
\toprule
\multirow{2}{*}{Method} & \multicolumn{4}{c|}{ImageNet-LT} & \multicolumn{4}{c}{iNaturalist2018} \\
\cmidrule(lr){2-5} \cmidrule(lr){6-9}
& Many & Med.
 & Few & All & Many & Med. & Few & All \\
\midrule
CE & 65.9 & 37.5 & 7.7 & 44.4 & 72.2 & 63.0 & 57.2 & 61.7 \\
$\tau$-norm \cite{09217} & 59.1 & 46.9 & 30.7 & 49.4 & 65.6 & 65.3 & 65.5 & 65.6 \\
cRT \cite{09217} & 61.8 & 46.2 & 27.3 & 49.6 & 69.0 & 66.0 & 63.2 & 65.2 \\
AREA \cite{Chen_2023_ICCV}  & - & - & - & 49.5 & - & - & - & 68.4 \\
LogitAdjust \cite{menon2021longtail} & - & - & - & 51.1 & - & - & - & 69.4 \\
MARC \cite{pmlr-v189-wang23b} & 60.4 
 & 50.3 & 36.6 & 52.3 & - & - & - & 70.4 \\
DiVE \cite{He_2021_ICCV} & 64.1 & 50.4 & 31.5 & 53.1 & 70.6 & 70.0 & 67.6 & 69.1 \\
EWB-FDR \cite{hasegawa2024exploringweightbalancinglongtailed}  & 63.4 & 50.0 & 35.1 & 53.2 & - & - & - & - \\
RBL \cite{pmlr-v202-peifeng23a}  & 64.8 & 49.6 & 34.2 & 53.3 & - & - & - & - \\
WD + MaxNorm \cite{Alshammari_2022_CVPR}
 & 62.5 & 50.4 & \textbf{41.5} & 53.9 & 71.2 & 70.4 & 69.7 & 70.2 \\
IWB \cite{Dang_Yang_Dong_Li_Shi_2024} & 64.2 & 52.2 & 40.2 & 55.2 & \textbf{72.3} & 70.6 & 72.5 & 71.5 \\
GLMC \cite{Du_2023_CVPR} & \textbf{70.1} & 52.4 & 30.4 & 56.3 & 64.6 & 73.2 & 73.0 & 72.2 \\
\midrule
\rowcolor{gray!20} GLMC + SAMN (Ours) & 69.7 & \textbf{56.1} & 31.1 & \textbf{57.7} & 64.4 & \textbf{73.7} & \textbf{73.6} & \textbf{72.7}   \\
\bottomrule
\end{tabular}
}
\begin{tablenotes}
      \scriptsize
      \item Note:``Many'' (class count $>$ 100), ``Medium'' (20 $\le$ class count $\le$ 100), and ``Few'' (class count $<$ 20)
    \end{tablenotes}
\end{table}

Two patterns are further observed in \cref{tab:1}.
Firstly, we observe that the effectiveness of SAMN is generally proportional to the dataset's imbalance factor (IF). The greatest accuracy improvements occur in datasets with higher IF values, revealing that SAMN is particularly adept at handling highly imbalanced data distributions.
Secondly, the performance improvements provided by SAMN in CIFAR-100-LT are consistently larger than those of CIFAR-10-LT. Since CIFAR-100 is recognized as a more challenging, fine-grained classification task than CIFAR-10 \cite{10.5555/1953048.2021069}, this indicates that SAMN excels at improving performance on more complex and difficult datasets.
These two properties demonstrate the immense value and robustness of the SAMN approach in real-world long-tailed scenarios. Besides, although we observe that ``Many'' class accuracy drops slightly in \cref{tab:2}, this is outweighed by larger gains in ``Medium'' and ``Few'' classes, proving SAMN's effectiveness. We hope to further mitigate this trade-off in the future by introducing some strategies of data augmentation.

The computational cost of SAMN has been shown in \cref{tab:overhead}. As can be seen, the overhead scales inversely with the computational load. Although visible in lightweight models like ResNet32 on CIFAR100 (12.8\%), it becomes negligible (2.9\%) in large-scale benchmarks like ResNeXt50 on ImageNet, where forward and backward passes dominate. This confirms SAMN's efficiency and scalability.

\begin{table}[t]
    \centering
    \small
    \setlength{\tabcolsep}{4pt}
    \caption{Computational overhead per epoch. } 
    \vspace{-0.2cm}
    \scalebox{0.95}{
    \begin{tabular}{c c c c c}
        \toprule
         \textbf{Dataset} & \textbf{Classes} & \textbf{Model} & \textbf{CE} & \textbf{CE + SAMN} \\
        \midrule
        CIFAR100 & 100 & ResNet32 & 3.9 s & 4.4 s ($\uparrow$ 12.8\%) \\
        ImageNet-LT & 1,000 & ResNeXt50 & 342.2 s & 352.0 s ($\uparrow$ 2.9\%) \\
        \bottomrule
    \end{tabular}
    }
    \label{tab:overhead}
\end{table}

\subsection{Ablation Studies}
\label{as}
This section evaluates key design choices in SAMN through two ablation studies. First, we investigate applying SAMN to classifier weights versus biases for rare class compensation. Second, we compare two order metrics for PAVA: class frequency and first-stage weight norms.

\noindent\textbf{Experimental Setup} We adopt the ResNet32 \cite{He_2016_CVPR} backbone and train it on CIFAR10-LT and CIFAR100-LT.
All setups are the same as those in \cref{cs} when using the basic CE method.
Meanwhile, we also report the results of the CE method as a baseline to compare with.

\noindent\textbf{Results} The results are shown in \cref{tab:3}. As can be seen, regardless of whether PAVA is applied to weights or biases, SAMN significantly improves the performance of the baseline. 
We find that the largest gap from the bias is 3.2\%, which appears in CIFAR100-LT with IF=100, while the largest gap from the weight is 6.6\%, appearing in the same situation.
It indicates that adjusting the weight is more significant than adjusting the bias through SAMN.
Crucially, utilizing SAMN for both weight and bias simultaneously produces the best results across all scenarios tested. This combined approach provided an additional increase of 0.6\% compared to using either alone (CIFAR10-LT, IF=100). Therefore, we strongly recommend that SAMN be initially applied to both the classifier weights and biases.

\begin{table}[!t]
\centering
\caption{Accuracy (\%) of Ablation studies on CIFAR100-LT and CIFAR10-LT datasets. \textbf{A}: order metric.
The imbalance factor is set to 100, 50, and 10. The best results are in \textbf{bold}.}
\label{tab:3}
\scalebox{0.61}
{
\begin{tabular}{c c cc ccc ccc}
\toprule
\multirow{2}{*}{Method} & \multirow{2}{*}{Order Metric ($\mathbf{A}$)} & \multicolumn{2}{c}{Components} & \multicolumn{3}{c}{CIFAR100-LT} & \multicolumn{3}{c}{CIFAR10-LT} \\
\cmidrule(lr){3-4} \cmidrule(lr){5-7} \cmidrule(lr){8-10} 
 & & Weight & Bias & 100 & 50 & 10 & 100 & 50 & 10 \\
\midrule


Baseline & - & - & - & 47.4 & 52.3 & 67.2 & 80.0 & 84.3 & 91.9 \\
\midrule

\multirow{6}{*}{SAMN} & \multirow{3}{*}{Class frequency} & - & \checkmark & 50.6 & 55.2 & 69.1 & 81.7 & 86.0 & 92.5 \\
 & & \checkmark & - & 53.9 & 58.3 & 70.2 & 83.6 & 87.3 & 92.8 \\
 & & \checkmark & \checkmark & \textbf{54.0} & \textbf{58.4} & 70.2 & \textbf{84.1} & 87.6 & \textbf{92.8} \\
\cmidrule(l){2-10} 

 & \multirow{3}{*}{Norms in stage one} & - & \checkmark & 50.1 & 55.0 & 69.0 & 81.9 & 86.2 & 92.2 \\
 & & \checkmark & - & 53.7 & 58.0 & 70.3 & 83.5 & 87.1 & 92.7 \\
 & & \checkmark & \checkmark & 53.9 & 58.2 & \textbf{70.3} & 83.9 & \textbf{87.7} & 92.7 \\
\bottomrule
\end{tabular}
}
\end{table}


Regarding the selection of the order metric \(\mathbf{A}\), we observe that using the class frequency and the learned norms obtained in the first stage yields similar overall performance. However, we maintain this selection as a flexible option for two primary reasons: First, the optimal results in \cref{tab:3} are distributed across both metrics, suggesting situational advantages. For example, using class frequency as the order metric achieves a 0.5\% improvement (50.6\% vs. 50.1\%) when applying SAMN only to the bias on CIFAR100-LT with IF=100. Conversely, applying the first-stage norms as the order metric achieves a 0.2\% improvement (81.9\% vs. 81.7\% and 86.2\% vs. 86.0\%) when applying SAMN only to the bias on CIFAR10-LT with IF=10 and IF=100, respectively. Second, the two metrics are motivated by different rationales. The class frequency represents the general static tendency of the required compensation for a given task, while the obtained norms in the first stage capture the specific dynamic situation of a particular training process.
\section{Conclusions}
In this paper, we focus on the two-stage decoupling strategy for long-tailed recognition and specifically explore norm-rescaling methods in the classifier retraining stage.
We first provide insight into the class-conditional distribution, demonstrating theoretically and empirically that performance issues in rare classes stem from underfitting rather than overfitting. This analysis strengthens the foundation for norm-rescaling strategies. We then identify a critical limitation in current adaptive norm rescaling approaches: they rely on parameter regularization, introducing sensitive hyperparameters that require careful tuning. To address this, we propose Self-Adaptive Monotonic Normalization, a simple, effective, and hyperparameter-friendly technique. It avoids parameter regularization by directly enforcing monotonicity on the per-class weight norms using the Pool Adjacent Violators Algorithm. Our method is a universal strategy that integrates with other state-of-the-art methods, achieving robust and superior performance.

\section*{Acknowledgments}
Tingting Zhu was supported by the Royal Academy of Engineering under the Research Fellowship scheme.

\ifarxiv \clearpage \appendix \maketitlesupplementary

The supplementary material provides additional derivations, discussions, and results to support the main paper. It is organized as follows: \cref{sec:Derivations} presents a detailed derivation for \cref{eq:10} and \cref{eq:11} from \cref{ou}; \cref{data} introduces the specific datasets and evaluation metrics used in this paper; \cref{sec:setup} describes the detailed experimental setup corresponding to \cref{cs}.

\section{Derivations of Class-Conditional Distributions}
\label{sec:Derivations}
According to \cref{eq:7}, the relationship between the model's logits and the class-conditional distribution $\mathcal{D}_{\mathrm{x}}^{k}(\mathbf{x}_i)$ is defined as:
\begin{equation}
    e^{w_k^\top \mathbf{x}_i + b_k} = Z_k \mathcal{D}_{\mathrm{x}}^{k}(\mathbf{x}_i),
    \label{eq:start}
\end{equation}
where $Z_k$ is the partition function (normalization constant).
Taking the natural logarithm of both sides of Eq. \eqref{eq:start} gives:
\begin{equation}
    w_k^\top \mathbf{x}_i + b_k = \log \mathcal{D}_{\mathrm{x}}^{k}(\mathbf{x}_i) + \log Z_k.
    \label{eq:log}
\end{equation}
Next, we multiply by a factor \(\alpha \) where \(0<\alpha <1\):
\begin{equation}
    \alpha w_k^\top \mathbf{x}_i + \alpha b_k = \alpha \log \mathcal{D}_{\mathrm{x}}^{k}(\mathbf{x}_i) + \alpha \log Z_k.
    \label{eq:alpha}
\end{equation}
We then add $(1-\alpha)b_k$ to both sides of \cref{eq:alpha}. Rearranging the terms and simplifying the left-hand side yields:
\begin{equation}
\begin{split}
    \alpha w_k^\top \mathbf{x}_i + b_k 
    &= (\alpha w_k^\top \mathbf{x}_i + \alpha b_k) + (1-\alpha)b_k \\
    &= \alpha \log \mathcal{D}_{\mathrm{x}}^{k}(\mathbf{x}_i) + \alpha \log Z_k \\
    & \quad + (1-\alpha)b_k.
\end{split}
\label{eq:rearrange}
\end{equation}
Now, we exponentiate both sides of Eq. \eqref{eq:rearrange}:
\begin{equation}
\begin{split}
    e^{\alpha w_k^\top \mathbf{x}_i + b_k} 
    &= \exp ( \alpha \log \mathcal{D}_{\mathrm{x}}^{k}(\mathbf{x}_i) + \alpha \log Z_k \\
    & \quad + (1-\alpha)b_k) \\ 
    &= \exp(\alpha \log \mathcal{D}_{\mathrm{x}}^{k}(\mathbf{x}_i)) \cdot \exp(\alpha \log Z_k) \\
    & \quad \cdot \exp((1-\alpha)b_k) \\
    &= [\mathcal{D}_{\mathrm{x}}^{k}(\mathbf{x}_i)]^\alpha (Z_k)^\alpha e^{(1-\alpha)b_k}.
\end{split}
\label{eq:exp}
\end{equation}
We define our new, unnormalized distribution, $\mathcal{D'}_{\mathrm{x}}^{k}(\mathbf{x}_i)$, as being proportional to the new exponentiated logits $e^{\alpha w_k^\top \mathbf{x}_i + b_k}$. From Eq. \eqref{eq:exp}, we have:
\begin{equation}
\begin{split}
    \mathcal{D'}_{\mathrm{x}}^{k}(\mathbf{x}_i) &\propto e^{\alpha w_k^\top \mathbf{x}_i + b_k} \\
            &\propto [\mathcal{D}_{\mathrm{x}}^{k}(\mathbf{x}_i)]^\alpha (Z_k)^\alpha e^{(1-\alpha)b_k}.
\end{split}
\label{eq:prop}
\end{equation}
This unnormalized relationship corresponds to \cref{eq:10} in the main paper. 

Since both $(Z_k)^\alpha$ and $e^{(1-\alpha)b_k}$ are independent of $\mathbf{x}_i$, they can be omitted as constant factors, yielding the proportional relationship:
\begin{equation}
    \mathcal{D'}_{\mathrm{x}}^{k}(\mathbf{x}_i) \propto [\mathcal{D}_{\mathrm{x}}^{k}(\mathbf{x}_i)]^\alpha.
    \label{eq:prop_simple}
\end{equation}
Finally, we normalize the distribution. We introduce a normalization constant $C'$ such that the distribution integrates to one:
\begin{equation}
    \mathcal{D'}_{\mathrm{x}}^{k}(\mathbf{x}_i) = C' \cdot [\mathcal{D}_{\mathrm{x}}^{k}(\mathbf{x}_i)]^\alpha.
\end{equation}
We solve for $C'$ by integrating over the entire domain (using $\mathbf{x}$ as the integration variable):
\begin{equation}
\begin{aligned}
\int \mathcal{D'}_{\mathrm{x}}^{k}(\mathbf{x}) \, d\mathbf{x} &= 1 \\
\int C' \cdot [\mathcal{D}_{\mathrm{x}}^{k}(\mathbf{x})]^\alpha \, d\mathbf{x} &= 1\\
C' \int [\mathcal{D}_{\mathrm{x}}^{k}(\mathbf{x})]^\alpha \, d\mathbf{x} &= 1.
\end{aligned}
\end{equation}
This gives us the constant:
\begin{equation}
    C' = \frac{1}{\int [\mathcal{D}_{\mathrm{x}}^{k}(\mathbf{x})]^\alpha \, d\mathbf{x}}
\end{equation}
Substituting $C'$ back into our expression for $\mathcal{D}_{\mathrm{x}}^{'k}(\mathbf{x}_i)$ yields the final, normalized form of \cref{eq:11} in the main paper:
\begin{equation}
    \mathcal{D'}_{\mathrm{x}}^{k}(\mathbf{x}_i) = \frac{[\mathcal{D}_{\mathrm{x}}^{k}(\mathbf{x}_i)]^\alpha}{\int [\mathcal{D}_{\mathrm{x}}^{k}(\mathbf{x})]^\alpha \, d\mathbf{x}}
    \label{eq:final}
\end{equation}
As this derivation shows, all constant factors from Eq. \eqref{eq:exp}, namely $(Z_k)^\alpha$ and $e^{(1-\alpha)b_k}$, are constants with respect to $\mathbf{x}_i$ and are absorbed into the normalization constant $C'$, ultimately canceling out during normalization. The integration variable $\mathbf{x}$ in the denominator is a bound variable, distinct from the specific instance $\mathbf{x}_i$.

\section{Datasets and Evaluation Metrics}
\label{data}
In this paper, we use four datasets: CIFAR10-LT \cite{10.5555/1953048.2021069}, CIFAR100-LT \cite{10.5555/1953048.2021069}, ImageNet-LT \cite{Liu_2019_CVPR}, iNaturalist2018 \cite{Horn_2018_CVPR} to perform the experiments.
CIFAR10-LT and CIFAR100-LT are the long-tailed versions of CIFAR10 and CIFAR100.
They are generated by downsampling the per-class training samples using the exponential decay function.
The degree of imbalance is controlled by the imbalance factor (IF) mentioned in \cref{pre}.
We set IF $\in$ \{10, 50, 100\} in this paper, while their validation sets are still balanced.
ImageNet-LT was introduced in \cite{Liu_2019_CVPR} by artificially truncating the original ImageNet dataset \cite{5206848}.
It has 1,000 classes, and the number of per-class training data ranges from 5 to 1280, with an IF of 256. iNaturalist2018 is another real-world dataset for the identification of species of animals and plants, containing 8,142 classes with an extremely large IF of 500. Regarding the evaluation metric, we use the overall accuracy in all datasets.
In addition, we also classify the classes into three subsets of ``Many'' ($n_k>$ 100), ``Medium'' (20 $\le n_k$ $\le$ 100), and ``Few'' ( $n_k<$ 20) and report their accuracy on ImageNet-LT and iNaturalist2018.

\section{Detailed Setup in \cref{cs}}
\label{sec:setup}

We evaluate SAMN on the benchmark datasets CIFAR10-LT and CIFAR100-LT, as well as the large-scale datasets ImageNet-LT and iNaturalist2018. All experiments utilize a stochastic gradient descent (SGD) optimizer with 0.9 momentum and a cosine learning rate scheduler \cite{DBLP:journals/corr/LoshchilovH16a}. For the CIFAR10-LT and CIFAR100-LT datasets, we adopt the ResNet32 \cite{He_2016_CVPR} backbone and apply SAMN to models pre-trained with CE, GLMC \cite{Du_2023_CVPR}, and SLAS \cite{Zhong_2021_CVPR}. The first stage training for CE and GLMC runs for 200 epochs with a batch size of 64, a weight decay of 5e-3, and an initial learning rate of 0.01. The hyperparameters of GLMC and SLAS follow their original implementations. In the second stage  (applying SAMN), we retrain only the classifiers for 20 epochs (batch size = 64), using an initial learning rate of 5e-4 for CE, 5e-5 for GLMC, and 1e-5 (CIFAR10-LT) or 5e-4 (CIFAR100-LT) for SLAS. For the large-scale datasets, we use a ResNeXt50 \cite{Xie_2017_CVPR} backbone, employing GLMC for first-stage training and SAMN for second-stage enhancement. Following \cite{Du_2023_CVPR}, on ImageNet-LT, the first stage is trained for 135 epochs (batch size = 128, weight decay = 2e-4, learning rate = 0.1), and the second stage is finetuned for 20 epochs (batch size = 512, learning rate = 1e-5). On iNaturalist2018, the first stage is trained for 120 epochs (batch size = 128, weight decay = 5e-3, learning rate = 0.1), and the second stage is finetuned for 20 epochs (batch size = 512, learning rate = 5e-5). Notably, weight decay was not applied during the second stage in any experiment.
 \fi


{
    \small
    \begin{thebibliography}{50}
\providecommand{\natexlab}[1]{#1}
\providecommand{\url}[1]{\texttt{#1}}
\expandafter\ifx\csname urlstyle\endcsname\relax
  \providecommand{\doi}[1]{doi: #1}\else
  \providecommand{\doi}{doi: \begingroup \urlstyle{rm}\Url}\fi

\bibitem[Ahn et~al.(2023)Ahn, Ko, and Yun]{ahn2023cuda}
Sumyeong Ahn, Jongwoo Ko, and Se-Young Yun.
\newblock {CUDA}: Curriculum of data augmentation for long-tailed recognition.
\newblock In \emph{The Eleventh International Conference on Learning Representations}, 2023.

\bibitem[Alshammari et~al.(2022)Alshammari, Wang, Ramanan, and Kong]{Alshammari_2022_CVPR}
Shaden Alshammari, Yu-Xiong Wang, Deva Ramanan, and Shu Kong.
\newblock Long-tailed recognition via weight balancing.
\newblock In \emph{Proceedings of the IEEE/CVF Conference on Computer Vision and Pattern Recognition (CVPR)}, 2022.

\bibitem[Ando and Huang(2017)]{10.1007/978-3-319-71249-9_46}
Shin Ando and Chun~Yuan Huang.
\newblock Deep over-sampling framework for classifying imbalanced data.
\newblock In \emph{Machine Learning and Knowledge Discovery in Databases}, 2017.

\bibitem[Ayer et~al.(1955)Ayer, Brunk, Ewing, Reid, and Silverman]{ff03617f-3b30-3cb3-a0fc-02979e761de5}
Miriam Ayer, H.~D. Brunk, G.~M. Ewing, W.~T. Reid, and Edward Silverman.
\newblock An empirical distribution function for sampling with incomplete information.
\newblock \emph{The Annals of Mathematical Statistics}, 26\penalty0 (4):\penalty0 641--647, 1955.

\bibitem[Cao et~al.(2019)Cao, Wei, Gaidon, Arechiga, and Ma]{NEURIPS2019_621461af}
Kaidi Cao, Colin Wei, Adrien Gaidon, Nikos Arechiga, and Tengyu Ma.
\newblock Learning imbalanced datasets with label-distribution-aware margin loss.
\newblock In \emph{Advances in Neural Information Processing Systems}, 2019.

\bibitem[Chawla et~al.(2002)Chawla, Bowyer, Hall, and Kegelmeyer]{10.5555/1622407.1622416}
Nitesh~V. Chawla, Kevin~W. Bowyer, Lawrence~O. Hall, and W.~Philip Kegelmeyer.
\newblock Smote: synthetic minority over-sampling technique.
\newblock \emph{Journal of Artificial Intelligence Research}, 16\penalty0 (1):\penalty0 321–357, 2002.

\bibitem[Chen et~al.(2023)Chen, Zhou, Wu, Yang, Li, Hu, and Wang]{Chen_2023_ICCV}
Xiaohua Chen, Yucan Zhou, Dayan Wu, Chule Yang, Bo Li, Qinghua Hu, and Weiping Wang.
\newblock Area: Adaptive reweighting via effective area for long-tailed classification.
\newblock In \emph{Proceedings of the IEEE/CVF International Conference on Computer Vision (ICCV)}, 2023.

\bibitem[Chu et~al.(2020)Chu, Bian, Liu, and Ling]{10.1007/978-3-030-58526-6_41}
Peng Chu, Xiao Bian, Shaopeng Liu, and Haibin Ling.
\newblock Feature space augmentation for long-tailed data.
\newblock \emph{arXiv preprint arXiv:2008.03673}, 2020.

\bibitem[Cui et~al.(2019)Cui, Jia, Lin, Song, and Belongie]{Cui_2019_CVPR}
Yin Cui, Menglin Jia, Tsung-Yi Lin, Yang Song, and Serge Belongie.
\newblock Class-balanced loss based on effective number of samples.
\newblock In \emph{Proceedings of the IEEE/CVF Conference on Computer Vision and Pattern Recognition (CVPR)}, 2019.

\bibitem[Dang et~al.(2024)Dang, Yang, Dong, Li, and Shi]{Dang_Yang_Dong_Li_Shi_2024}
Wenqi Dang, Zhou Yang, Weisheng Dong, Xin Li, and Guangming Shi.
\newblock Inverse weight-balancing for deep long-tailed learning.
\newblock In \emph{Proceedings of the AAAI Conference on Artificial Intelligence}, 2024.

\bibitem[Deng et~al.(2009)Deng, Dong, Socher, Li, Li, and Fei-Fei]{5206848}
Jia Deng, Wei Dong, Richard Socher, Li-Jia Li, Kai Li, and Li Fei-Fei.
\newblock Imagenet: A large-scale hierarchical image database.
\newblock In \emph{2009 IEEE Conference on Computer Vision and Pattern Recognition}, pages 248--255, 2009.

\bibitem[Du et~al.(2023)Du, Yang, Jia, Nan, Chen, and Yang]{Du_2023_CVPR}
Fei Du, Peng Yang, Qi Jia, Fengtao Nan, Xiaoting Chen, and Yun Yang.
\newblock Global and local mixture consistency cumulative learning for long-tailed visual recognitions.
\newblock In \emph{Proceedings of the IEEE/CVF Conference on Computer Vision and Pattern Recognition (CVPR)}, 2023.

\bibitem[Duchi et~al.(2011)Duchi, Hazan, and Singer]{10.5555/1953048.2021068}
John Duchi, Elad Hazan, and Yoram Singer.
\newblock Adaptive subgradient methods for online learning and stochastic optimization.
\newblock \emph{The Journal of Machine Learning Research}, page 2121–2159, 2011.

\bibitem[Hanson and Pratt(1988)]{NIPS1988_1c9ac015}
Stephen Hanson and Lorien Pratt.
\newblock Comparing biases for minimal network construction with back-propagation.
\newblock In \emph{Advances in Neural Information Processing Systems}, 1988.

\bibitem[Hasegawa and Sato(2024)]{hasegawa2024exploringweightbalancinglongtailed}
Naoya Hasegawa and Issei Sato.
\newblock Exploring weight balancing on long-tailed recognition problem.
\newblock \emph{arXiv preprint arXiv:2305.16573}, 2024.

\bibitem[He et~al.(2016)He, Zhang, Ren, and Sun]{He_2016_CVPR}
Kaiming He, Xiangyu Zhang, Shaoqing Ren, and Jian Sun.
\newblock Deep residual learning for image recognition.
\newblock In \emph{Proceedings of the IEEE Conference on Computer Vision and Pattern Recognition (CVPR)}, 2016.

\bibitem[He et~al.(2021)He, Wu, and Wei]{He_2021_ICCV}
Yin-Yin He, Jianxin Wu, and Xiu-Shen Wei.
\newblock Distilling virtual examples for long-tailed recognition.
\newblock In \emph{Proceedings of the IEEE/CVF International Conference on Computer Vision (ICCV)}, 2021.

\bibitem[Helal et~al.(2016)Helal, Haydar, and Mostafa]{7860349}
Mustakim~Al Helal, Mohammad~Salman Haydar, and Seraj Al~Mahmud Mostafa.
\newblock Algorithms efficiency measurement on imbalanced data using geometric mean and cross validation.
\newblock In \emph{2016 International Workshop on Computational Intelligence (IWCI)}, 2016.

\bibitem[Kang et~al.(2019)Kang, Xie, Rohrbach, Yan, Gordo, Feng, and Kalantidis]{09217}
Bingyi Kang, Saining Xie, Marcus Rohrbach, Zhicheng Yan, Albert Gordo, Jiashi Feng, and Yannis Kalantidis.
\newblock Decoupling representation and classifier for long-tailed recognition.
\newblock \emph{arXiv preprint arXiv:1910.09217}, 2019.

\bibitem[Kim and Kim(2020)]{9081988}
Byungju Kim and Junmo Kim.
\newblock Adjusting decision boundary for class imbalanced learning.
\newblock \emph{IEEE Access}, 8:\penalty0 81674--81685, 2020.

\bibitem[Krizhevsky and Hinton(2009)]{10.5555/1953048.2021069}
Alex Krizhevsky and Geoffrey Hinton.
\newblock Learning multiple layers of features from tiny images.
\newblock \emph{Master's thesis, University of Tront}, 2009.

\bibitem[Lema{{\^i}}tre et~al.(2017)Lema{{\^i}}tre, Nogueira, and Aridas]{JMLR:v18:16-365}
Guillaume Lema{{\^i}}tre, Fernando Nogueira, and Christos~K. Aridas.
\newblock Imbalanced-learn: A python toolbox to tackle the curse of imbalanced datasets in machine learning.
\newblock \emph{Journal of Machine Learning Research}, 18\penalty0 (17):\penalty0 1--5, 2017.

\bibitem[Li et~al.(2016)Li, Fong, Mohammed, Fiaidhi, Chen, and Tan]{10105459}
Jinyan Li, Simon Fong, Sabah Mohammed, Jinan Fiaidhi, Qian Chen, and Zhen Tan.
\newblock Solving the under-fitting problem for decision tree algorithms by incremental swarm optimization in rare-event healthcare classification.
\newblock \emph{Journal of Medical Imaging and Health Informatics}, 6\penalty0 (4):\penalty0 1102--1110, 2016.

\bibitem[Li et~al.(2018)Li, Fong, Wong, and Chu]{LI20181}
Jinyan Li, Simon Fong, Raymond~K. Wong, and Victor~W. Chu.
\newblock Adaptive multi-objective swarm fusion for imbalanced data classification.
\newblock \emph{Information Fusion}, 39:\penalty0 1--24, 2018.

\bibitem[Li et~al.(2022)Li, Cheung, and Lu]{Li_2022_CVPR}
Mengke Li, Yiu-ming Cheung, and Yang Lu.
\newblock Long-tailed visual recognition via gaussian clouded logit adjustment.
\newblock In \emph{Proceedings of the IEEE/CVF Conference on Computer Vision and Pattern Recognition (CVPR)}, 2022.

\bibitem[Li et~al.(2021)Li, Gong, Liu, Wang, Qiao, and Cheng]{Li_2021_CVPR}
Shuang Li, Kaixiong Gong, Chi~Harold Liu, Yulin Wang, Feng Qiao, and Xinjing Cheng.
\newblock Metasaug: Meta semantic augmentation for long-tailed visual recognition.
\newblock In \emph{Proceedings of the IEEE/CVF Conference on Computer Vision and Pattern Recognition (CVPR)}, 2021.

\bibitem[Lin et~al.(2017)Lin, Goyal, Girshick, He, and Dollar]{Lin_2017_ICCV}
Tsung-Yi Lin, Priya Goyal, Ross Girshick, Kaiming He, and Piotr Dollar.
\newblock Focal loss for dense object detection.
\newblock In \emph{Proceedings of the IEEE International Conference on Computer Vision (ICCV)}, 2017.

\bibitem[Liu et~al.(2009)Liu, Wu, and Zhou]{4717268}
Xu-Ying Liu, Jianxin Wu, and Zhi-Hua Zhou.
\newblock Exploratory undersampling for class-imbalance learning.
\newblock \emph{IEEE Transactions on Systems, Man, and Cybernetics, Part B (Cybernetics)}, 39\penalty0 (2):\penalty0 539--550, 2009.

\bibitem[Liu et~al.(2019)Liu, Miao, Zhan, Wang, Gong, and Yu]{Liu_2019_CVPR}
Ziwei Liu, Zhongqi Miao, Xiaohang Zhan, Jiayun Wang, Boqing Gong, and Stella~X. Yu.
\newblock Large-scale long-tailed recognition in an open world.
\newblock In \emph{Proceedings of the IEEE/CVF Conference on Computer Vision and Pattern Recognition (CVPR)}, 2019.

\bibitem[Loshchilov and Hutter(2016)]{DBLP:journals/corr/LoshchilovH16a}
Ilya Loshchilov and Frank Hutter.
\newblock {SGDR:} stochastic gradient descent with restarts.
\newblock \emph{arXiv preprint arXiv:1608.03983}, 2016.

\bibitem[Menon et~al.(2021)Menon, Jayasumana, Rawat, Jain, Veit, and Kumar]{menon2021longtail}
Aditya~Krishna Menon, Sadeep Jayasumana, Ankit~Singh Rawat, Himanshu Jain, Andreas Veit, and Sanjiv Kumar.
\newblock Long-tail learning via logit adjustment.
\newblock In \emph{International Conference on Learning Representations}, 2021.

\bibitem[Park et~al.(2022)Park, Hong, Heo, Yun, and Choi]{Park_2022_CVPR}
Seulki Park, Youngkyu Hong, Byeongho Heo, Sangdoo Yun, and Jin~Young Choi.
\newblock The majority can help the minority: Context-rich minority oversampling for long-tailed classification.
\newblock In \emph{Proceedings of the IEEE/CVF Conference on Computer Vision and Pattern Recognition (CVPR)}, 2022.

\bibitem[Peifeng et~al.(2023)Peifeng, Xu, Wen, Yang, Shao, and Huang]{pmlr-v202-peifeng23a}
Gao Peifeng, Qianqian Xu, Peisong Wen, Zhiyong Yang, Huiyang Shao, and Qingming Huang.
\newblock Feature directions matter: Long-tailed learning via rotated balanced representation.
\newblock In \emph{Proceedings of the 40th International Conference on Machine Learning}, 2023.

\bibitem[Rangwani et~al.(2022)Rangwani, Aithal, Mishra, and R]{NEURIPS2022_8f4d70db}
Harsh Rangwani, Sumukh~K Aithal, Mayank Mishra, and Venkatesh~Babu R.
\newblock Escaping saddle points for effective generalization on class-imbalanced data.
\newblock In \emph{Advances in Neural Information Processing Systems}, 2022.

\bibitem[Ren et~al.(2020)Ren, Yu, sheng, Ma, Zhao, Yi, and Li]{NEURIPS2020_2ba61cc3}
Jiawei Ren, Cunjun Yu, shunan sheng, Xiao Ma, Haiyu Zhao, Shuai Yi, and hongsheng Li.
\newblock Balanced meta-softmax for long-tailed visual recognition.
\newblock In \emph{Advances in Neural Information Processing Systems}, 2020.

\bibitem[Shao et~al.(2024)Shao, Zhu, Zhang, and Wu]{shao2024diffultmakediffusionmodel}
Jie Shao, Ke Zhu, Hanxiao Zhang, and Jianxin Wu.
\newblock Diffult: How to make diffusion model useful for long-tail recognition.
\newblock \emph{arXiv preprint arXiv:2403.05170}, 2024.

\bibitem[Shi et~al.(2023)Shi, Wei, Xiang, and Li]{NEURIPS2023_eeffa70b}
Jiang-Xin Shi, Tong Wei, Yuke Xiang, and Yu-Feng Li.
\newblock How re-sampling helps for long-tail learning?
\newblock In \emph{Advances in Neural Information Processing Systems}, 2023.

\bibitem[Van~Horn et~al.(2018)Van~Horn, Mac~Aodha, Song, Cui, Sun, Shepard, Adam, Perona, and Belongie]{Horn_2018_CVPR}
Grant Van~Horn, Oisin Mac~Aodha, Yang Song, Yin Cui, Chen Sun, Alex Shepard, Hartwig Adam, Pietro Perona, and Serge Belongie.
\newblock The inaturalist species classification and detection dataset.
\newblock In \emph{Proceedings of the IEEE Conference on Computer Vision and Pattern Recognition (CVPR)}, 2018.

\bibitem[Wang et~al.(2021)Wang, Lian, Miao, Liu, and Yu]{wang2021longtailed}
Xudong Wang, Long Lian, Zhongqi Miao, Ziwei Liu, and Stella Yu.
\newblock Long-tailed recognition by routing diverse distribution-aware experts.
\newblock In \emph{International Conference on Learning Representations}, 2021.

\bibitem[Wang et~al.(2023{\natexlab{a}})Wang, Zhang, Hou, Wu, Wang, and Shinozaki]{pmlr-v189-wang23b}
Yidong Wang, Bowen Zhang, Wenxin Hou, Zhen Wu, Jindong Wang, and Takahiro Shinozaki.
\newblock Margin calibration for long-tailed visual recognition.
\newblock In \emph{Proceedings of The 14th Asian Conference on Machine Learning}, 2023{\natexlab{a}}.

\bibitem[Wang et~al.(2023{\natexlab{b}})Wang, Xu, Yang, He, Cao, and Huang]{NEURIPS2023_973a0f50}
Zitai Wang, Qianqian Xu, Zhiyong Yang, Yuan He, Xiaochun Cao, and Qingming Huang.
\newblock A unified generalization analysis of re-weighting and logit-adjustment for imbalanced learning.
\newblock In \emph{Advances in Neural Information Processing Systems}, 2023{\natexlab{b}}.

\bibitem[Xie et~al.(2017)Xie, Girshick, Dollar, Tu, and He]{Xie_2017_CVPR}
Saining Xie, Ross Girshick, Piotr Dollar, Zhuowen Tu, and Kaiming He.
\newblock Aggregated residual transformations for deep neural networks.
\newblock In \emph{Proceedings of the IEEE Conference on Computer Vision and Pattern Recognition (CVPR)}, 2017.

\bibitem[Xu and Lyu(2024)]{10105458}
Yuge Xu and Chuanlong Lyu.
\newblock Class-balanced regularization for long-tailed recognition.
\newblock \emph{Neural Processing Letters}, 56\penalty0 (128):\penalty0 1--18, 2024.

\bibitem[Ye et~al.(2020)Ye, Chen, Zhan, and Chao]{DBLP:journals/corr/abs-2001-01385}
Han{-}Jia Ye, Hong{-}You Chen, De{-}Chuan Zhan, and Wei{-}Lun Chao.
\newblock Identifying and compensating for feature deviation in imbalanced deep learning.
\newblock \emph{arXiv preprint arXiv:2001.01385}, 2020.

\bibitem[Ye et~al.(2021)Ye, Zhan, and Chao]{Ye_2021_ICCV}
Han-Jia Ye, De-Chuan Zhan, and Wei-Lun Chao.
\newblock Procrustean training for imbalanced deep learning.
\newblock In \emph{Proceedings of the IEEE/CVF International Conference on Computer Vision (ICCV)}, 2021.

\bibitem[Zhang et~al.(2021)Zhang, Wei, Zhou, and Wu]{Zhang_Wei_Zhou_Wu_2021}
Yongshun Zhang, Xiu-Shen Wei, Boyan Zhou, and Jianxin Wu.
\newblock Bag of tricks for long-tailed visual recognition with deep convolutional neural networks.
\newblock In \emph{Proceedings of the AAAI Conference on Artificial Intelligence}, 2021.

\bibitem[Zhang et~al.(2022)Zhang, Hooi, Hong, and Feng]{NEURIPS2022_dc6319dd}
Yifan Zhang, Bryan Hooi, Lanqing Hong, and Jiashi Feng.
\newblock Self-supervised aggregation of diverse experts for test-agnostic long-tailed recognition.
\newblock In \emph{Advances in Neural Information Processing Systems}, 2022.

\bibitem[Zhang et~al.(2023)Zhang, Kang, Hooi, Yan, and Feng]{10105457}
Yifan Zhang, Bingyi Kang, Bryan Hooi, Shuicheng Yan, and Jiashi Feng.
\newblock Deep long-tailed learning: A survey.
\newblock \emph{IEEE Transactions on Pattern Analysis and Machine Intelligence}, 45\penalty0 (9):\penalty0 10795--10816, 2023.

\bibitem[Zhong et~al.(2021)Zhong, Cui, Liu, and Jia]{Zhong_2021_CVPR}
Zhisheng Zhong, Jiequan Cui, Shu Liu, and Jiaya Jia.
\newblock Improving calibration for long-tailed recognition.
\newblock In \emph{Proceedings of the IEEE/CVF Conference on Computer Vision and Pattern Recognition (CVPR)}, 2021.

\bibitem[Zhou et~al.(2020)Zhou, Cui, Wei, and Chen]{Zhou_2020_CVPR}
Boyan Zhou, Quan Cui, Xiu-Shen Wei, and Zhao-Min Chen.
\newblock Bbn: Bilateral-branch network with cumulative learning for long-tailed visual recognition.
\newblock In \emph{Proceedings of the IEEE/CVF Conference on Computer Vision and Pattern Recognition (CVPR)}, 2020.

\end{thebibliography}

}
\end{document}